%% file: 0_main.tex
\documentclass{article}
\usepackage{style/unites}
\usepackage{XCharter}
\usepackage[scaled=1.1]{zlmtt}

\input{style/macros}

\begin{document}

\makeatletter
\def\blfootnote{\gdef\@thefnmark{}\@footnotetext}
\makeatother

\makeatletter
\pagestyle{fancy}
\fancyhf{}
\renewcommand{\headrulewidth}{1pt}
\chead{\small\bf \input{1_title}
}
\cfoot{\thepage}
\thispagestyle{fancy}
\makeatother

\makeatletter
\def\icmldate#1{\gdef\@icmldate{#1}}
\icmldate{\today}
\makeatother

\makeatletter
\fancypagestyle{fancytitlepage}{
  \fancyhead{}
  \lhead{\includegraphics[height=0.8cm]{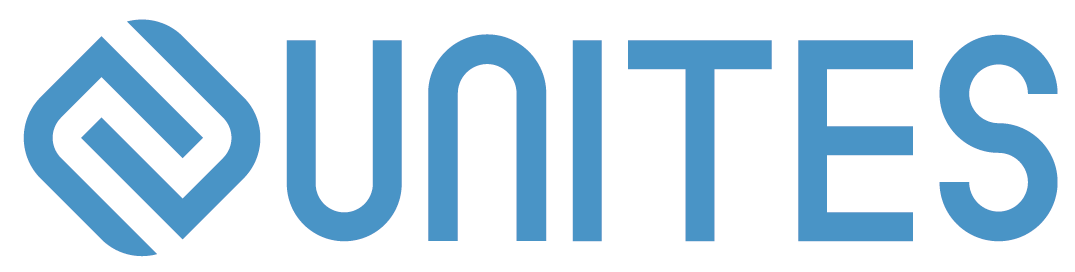}}
  \rhead{\it \@icmldate}
  \cfoot{}
}
\makeatother

\thispagestyle{fancytitlepage}

\vspace*{0.5em}

\noindent
\begin{titleblock}
    {\setlength{\parskip}{0cm}
     \raggedright
     {\setstretch{1.2}
      \LARGE\bfseries
      \input{1_title}
      \par}
    }
    \vskip 0.2cm
    
    \input{2_authors}
    \vskip 0.2cm
    
    \input{tex/0_abs}
    
    \vskip 0.2cm
    {\setlength{\parskip}{0cm}
     \centering
     \makebox[\linewidth]{
        \metadataformat[Code]{
            \href{https://github.com/UNITES-Lab/linear-vggt}{https://github.com/UNITES-Lab/linear-vggt}
        }
     }
    }
\end{titleblock}

\blfootnote{%
$^{\textrm{\Letter}}$ Corresponding authors: \{haoyuk, tianlong\}@cs.unc.edu
\\[2.5em]
\ifcsname @icmlpreprint\endcsname
  \textit{\csname @icmlpreprint\endcsname}%
\fi
}

\input{tex/1_intro}
\input{tex/3_method}
\input{tex/4_experiments}

\input{tex/2_related.tex}
\input{tex/7_conclusion}

\bibliography{999_reference}
\bibliographystyle{style/icml2025}

\titlespacing*{\section}{0pt}{*1}{*1}
\titlespacing*{\subsection}{0pt}{*1.25}{*1.25}
\titlespacing*{\subsubsection}{0pt}{*1.5}{*1.5}

\setlength{\abovedisplayskip}{\baselineskip}
\setlength{\abovedisplayshortskip}{0.5\baselineskip}
\setlength{\belowdisplayskip}{\baselineskip}
\setlength{\belowdisplayshortskip}{0.5\baselineskip}

\clearpage
\appendix
\label{sec:append}
\part*{Appendix}
{
\setlength{\parskip}{-0em}
\startcontents[sections]
\printcontents[sections]{ }{1}{}
}

\setlength{\parskip}{.5em}
\input{tex/5_appendix}

\end{document}

%% file: style/macros.tex
\usepackage[utf8]{inputenc}
\usepackage[T1]{fontenc}
\usepackage{microtype}

\usepackage{amsmath}
\usepackage{amssymb}
\usepackage{amsfonts}
\usepackage{amsthm}
\usepackage{mathtools}
\usepackage{mathrsfs}
\usepackage{physics}
\usepackage{braket}
\usepackage{slashed}
\usepackage{nicefrac}
\usepackage{textcomp}
\usepackage{dsfont}
\usepackage{bbm}
\usepackage{bm}

\usepackage{graphicx}
\usepackage{subcaption}
\usepackage[export]{adjustbox}
\usepackage{float}
\usepackage{booktabs}
\usepackage{dcolumn}
\newcolumntype{d}[1]{D{.}{.}{#1}}
\usepackage{bigstrut, tabularx, multirow, makecell, diagbox}
\usepackage{colortbl}
\usepackage{tabularray}
\UseTblrLibrary{booktabs}
\usepackage{threeparttable}
\usepackage{tablefootnote}
\usepackage{fontawesome5}

\usepackage{placeins}
\usepackage{caption}
\usepackage{footnote}
\usepackage{enumitem}
\usepackage{multicol}
\usepackage{xspace}
\usepackage{titletoc}
\usepackage{titlesec}
\usepackage[bottom]{footmisc}
\usepackage{setspace}

\usepackage{wrapfig}
\usepackage{tikz}
\usepackage{quantikz}
\usepackage{dashbox}
\usepackage{mdframed}
\usepackage{marvosym}
\usepackage{pifont}
\usepackage{CJK}
\usepackage{url}

\usepackage[table,x11names]{xcolor}
\usepackage[most]{tcolorbox}
\tcbuselibrary{breakable}
\usetikzlibrary{decorations.pathreplacing, fit}

\definecolor{primaryblue}{HTML}{0066CC}
\definecolor{accentcyan}{HTML}{00D4AA}
\definecolor{warmorange}{HTML}{FF6B35}
\definecolor{deepgray}{HTML}{2C3E50}
\definecolor{lightgray}{HTML}{F8F9FA}
\definecolor{gradientstart}{HTML}{667eea}
\definecolor{gradientend}{HTML}{764ba2}

\definecolor{citecolor}{HTML}{0071bc}
\definecolor{citeblue}{RGB}{0, 113, 188}
\definecolor{linkcolor}{HTML}{9A4D92}
\definecolor{firebrick}{rgb}{0.698,0.133,0.133}

\definecolor{paleviolet}{HTML}{E1EEFC}
\definecolor{CarolinaUltraLight}{HTML}{E7F4FC}
\definecolor{lightgrey}{RGB}{247, 247, 247}
\definecolor{shadecolor}{HTML}{EFEFEF}
\definecolor{lightyellow}{rgb}{1.0, 0.95, 0.7}
\definecolor{lightblue}{rgb}{0.90, 0.95, 1.0}
\definecolor{light-gray}{gray}{0.95}

\definecolor{darkgrey}{rgb}{0.5, 0.5, 0.5}
\definecolor{darkgreen}{rgb}{0, 0.5, 0}
\definecolor{mydarkblue}{rgb}{0,0.08,0.45}
\definecolor{mydarkblue2}{rgb}{0.133, 0.133, 0.698}
\definecolor{echodrk}{HTML}{0099cc}
\definecolor{mymauve}{rgb}{0.58,0,0.82}
\definecolor{midnightblue}{rgb}{0.1,0.1,0.44}
\definecolor{oxfordblue}{rgb}{0.0,0.13,0.28}
\definecolor{prussianblue}{rgb}{0.0,0.19,0.33}
\definecolor{coolteal}{rgb}{0, 0.45, 0.45}
\definecolor{olive}{rgb}{0.1, 0.3, 0}
\definecolor{mypurple}{rgb}{0.5,0,0.5}
\definecolor{almond}{rgb}{0.94, 0.87, 0.8}

\definecolor{blue_ampEncoding}{HTML}{DAE8FC}
\definecolor{green_encoder}{HTML}{D5E8D4}
\definecolor{purple_decoder}{HTML}{E1D5E7}
\definecolor{yellow_measure}{HTML}{FFF2CC}
\definecolor{gray_block}{HTML}{F5F5F5}
\definecolor{pink_dru}{HTML}{FAD9D5}
\definecolor{orange_v}{HTML}{FAD7AC}

\definecolor{colorA}{rgb}{1,0,0}
\definecolor{colorB}{rgb}{0,0.3,1}
\definecolor{colorC}{rgb}{0.9,0.8,0.2}
\definecolor{colorD}{rgb}{0,0.65,0}
\definecolor{lesslightgray}{rgb}{0.5,0.5,0.5}
\definecolor{fundamental}{RGB}{55, 110, 111}
\definecolor{Gred}{RGB}{219, 50, 54}
\definecolor{ToCgreen}{RGB}{0, 128, 0}
\definecolor{Sepia}{RGB}{112, 66, 20}
\definecolor{Dblue}{rgb}{0,0.08,0.45}
\definecolor{Blue}{rgb}{0, 0, 0.8}
\definecolor{blue}{rgb}{0,0,1}
\definecolor{UNCblue!10}{rgb}{0.84,0.91,0.98}
\definecolor{RowAlt}{rgb}{0.98,0.98,0.99}

\definecolor{CarolinaBlue}{HTML}{7BAFD4}        
\definecolor{CarolinaLightBlue}{HTML}{B3D4E5}   
\definecolor{CarolinaUltraLight}{HTML}{E8F4F8}  
\definecolor{CarolinaText}{HTML}{1C2B33}        

\usepackage[pagebackref=true,breaklinks=true,colorlinks,hyperfootnotes=false]{hyperref}
\hypersetup{
  colorlinks,
  citecolor=citeblue,
  linkcolor=firebrick,
  urlcolor=firebrick
}
\usepackage[nameinlink,capitalize,noabbrev]{cleveref}

\titlespacing\section{0pt}{4pt plus 4pt minus 2pt}{-2pt plus 2pt minus 2pt}
\titlespacing\subsection{0pt}{2pt plus 4pt minus 2pt}{-2pt plus 2pt minus 2pt}
\titlespacing\subsubsection{0pt}{2pt plus 4pt minus 2pt}{-2pt plus 2pt minus 2pt}

\makeatletter
\def\th@remark{%
  \thm@headfont{\bfseries}%
  \normalfont 
  \thm@preskip\topsep \divide\thm@preskip\tw@
  \thm@postskip\thm@preskip
}
\makeatother

\theoremstyle{definition}

\tcolorboxenvironment{theorem}{
  breakable,
  colback=black!10,
  colframe=white,
  width=\linewidth, 
  enlarge left by=0pt,
  enlarge right by=0pt,
  boxsep=5pt,
  boxrule=0pt,
  left=0pt,right=0pt,top=0pt,bottom=0pt,
  arc=8pt,
  before skip=\topsep,
  after skip=\topsep
}

\tcolorboxenvironment{lemma}{
  breakable,
  colback=black!10,
  colframe=white,
  width=\linewidth,
  enlarge left by=0pt,
  enlarge right by=0pt,
  boxsep=5pt,
  boxrule=0pt,
  left=0pt,right=0pt,top=0pt,bottom=0pt,
  arc=8pt,
  before skip=\topsep,
  after skip=\topsep
}

\tcolorboxenvironment{corollary}{
  breakable,
  colback=black!10,
  colframe=white,
  width=\linewidth,
  enlarge left by=0pt,
  enlarge right by=0pt,
  boxsep=5pt,
  boxrule=0pt,
  left=0pt,right=0pt,top=0pt,bottom=0pt,
  arc=8pt,
  before skip=\topsep,
  after skip=\topsep
}

\tcolorboxenvironment{proposition}{
  breakable,
  colback=black!10,
  colframe=white,
  width=\linewidth,
  enlarge left by=0pt,
  enlarge right by=0pt,
  boxsep=5pt,
  boxrule=0pt,
  left=0pt,right=0pt,top=0pt,bottom=0pt,
  arc=8pt,
  before skip=\topsep,
  after skip=\topsep
}

\tcolorboxenvironment{definition}{
  breakable,
  colback=black!10,
  colframe=white,
  width=\linewidth,
  enlarge left by=0pt,
  enlarge right by=0pt,
  boxsep=5pt,
  boxrule=0pt,
  left=0pt,right=0pt,top=0pt,bottom=0pt,
  arc=8pt,
  before skip=\topsep,
  after skip=\topsep
}

\tcolorboxenvironment{assumption}{
  breakable,
  colback=black!10,
  colframe=white,
  width=\linewidth,
  enlarge left by=0pt,
  enlarge right by=0pt,
  boxsep=5pt,
  boxrule=0pt,
  left=0pt,right=0pt,top=0pt,bottom=0pt,
  arc=8pt,
  before skip=\topsep,
  after skip=\topsep
}

\tcolorboxenvironment{claim}{
  breakable,
  colback=black!10,
  colframe=white,
  width=\linewidth,
  enlarge left by=0pt,
  enlarge right by=0pt,
  boxsep=5pt,
  boxrule=0pt,
  left=0pt,right=0pt,top=0pt,bottom=0pt,
  arc=8pt,
  before skip=\topsep,
  after skip=\topsep
}

\tcolorboxenvironment{problem}{
  breakable,
  colback=black!10,
  colframe=white,
  width=\linewidth,
  enlarge left by=0pt,
  enlarge right by=0pt,
  boxsep=5pt,
  boxrule=0pt,
  left=0pt,right=0pt,top=0pt,bottom=0pt,
  arc=8pt,
  before skip=\topsep,
  after skip=\topsep
}

\tcolorboxenvironment{question}{
  breakable,
  colback=black!10,
  colframe=white,
  width=\linewidth,
  enlarge left by=0pt,
  enlarge right by=0pt,
  boxsep=5pt,
  boxrule=0pt,
  left=0pt,right=0pt,top=0pt,bottom=0pt,
  arc=8pt,
  before skip=\topsep,
  after skip=\topsep
}

\newtcolorbox{titleblock}{
  enhanced,
  frame hidden,
  colback=CarolinaUltraLight,
  colframe=CarolinaUltraLight,
  boxrule=0pt,
  arc=10pt,
  left=14pt,
  right=14pt,
  top=14pt,
  bottom=14pt,
  width=\linewidth,
  before skip=12pt plus 4pt,
  after skip=12pt plus 4pt,
  grow to left by=1.5pt,
  grow to right by=1.5pt,
  before upper={
    \setlength{\parindent}{0cm}
    \setlength{\parskip}{0.5cm}
  }
}

\crefname{theorem}{Theorem}{Theorems}
\crefname{proposition}{Proposition}{Propositions}
\crefname{lemma}{Lemma}{Lemmas}
\crefname{corollary}{Corollary}{Corollaries}
\crefname{definition}{Definition}{Definitions}
\crefname{assumption}{Assumption}{Assumptions}
\crefname{remark}{Remark}{Remarks}
\crefname{problem}{Problem}{Problems}
\crefname{property}{Property}{property}
\crefname{question}{Question}{Questions}

\numberwithin{equation}{section}
\numberwithin{theorem}{section}
\numberwithin{proposition}{section}
\numberwithin{definition}{section}
\numberwithin{lemma}{section}
\numberwithin{assumption}{section}
\numberwithin{remark}{section}

\newcommand\metadataformat[2][]{{\small {\bfseries #1:} #2}}

\def\1{\bm{1}}

\makeatletter
\let\save@mathaccent\mathaccent
\newcommand*\if@single[3]{%
    \setbox0\hbox{${\mathaccent"0362{#1}}^H$}%
    \setbox2\hbox{${\mathaccent"0362{\kern0pt#1}}^H$}%
    \ifdim\ht0=\ht2 #3\else #2\fi
}
\newcommand*\rel@kern[1]{\kern#1\dimexpr\macc@kerna}
\newcommand*\widebar[1]{\@ifnextchar^{{\wide@bar{#1}{0}}}{\wide@bar{#1}{1}}}
\newcommand*\wide@bar[2]{\if@single{#1}{\wide@bar@{#1}{#2}{1}}{\wide@bar@{#1}{#2}{2}}}
\newcommand*\wide@bar@[3]{%
    \begingroup
    \def\mathaccent##1##2{%
        \let\mathaccent\save@mathaccent
        \if#32 \let\macc@nucleus\first@char \fi
        \setbox\z@\hbox{$\macc@style{\macc@nucleus}_{}$}%
        \setbox\tw@\hbox{$\macc@style{\macc@nucleus}{}_{}$}%
        \dimen@\wd\tw@
        \advance\dimen@-\wd\z@
        \divide\dimen@ 3
        \@tempdima\wd\tw@
        \advance\@tempdima-\scriptspace
        \divide\@tempdima 10
        \advance\dimen@-\@tempdima
        \ifdim\dimen@>\z@ \dimen@0pt\fi
        \rel@kern{0.6}\kern-\dimen@
        \if#31
        \overline{\rel@kern{-0.6}\kern\dimen@\macc@nucleus\rel@kern{0.4}\kern\dimen@}%
        \advance\dimen@0.4\dimexpr\macc@kerna
        \let\final@kern#2%
        \ifdim\dimen@<\z@ \let\final@kern1\fi
        \if\final@kern1 \kern-\dimen@\fi
        \else
        \overline{\rel@kern{-0.6}\kern\dimen@#1}%
        \fi
    }%
    \macc@depth\@ne
    \let\math@bgroup\@empty \let\math@egroup\macc@set@skewchar
    \mathsurround\z@ \frozen@everymath{\mathgroup\macc@group\relax}%
    \macc@set@skewchar\relax
    \let\mathaccentV\macc@nested@a
    \if#31
    \macc@nested@a\relax111{#1}%
    \else
    \def\gobble@till@marker##1\endmarker{}%
    \futurelet\first@char\gobble@till@marker#1\endmarker
    \ifcat\noexpand\first@char A\else
    \def\first@char{}%
    \fi
    \macc@nested@a\relax111{\first@char}%
    \fi
    \endgroup
    }
\makeatother
\let\bar\widebar

\DeclareMathAlphabet{\mathsfit}{\encodingdefault}{\sfdefault}{m}{sl}
\SetMathAlphabet{\mathsfit}{bold}{\encodingdefault}{\sfdefault}{bx}{n}

\let\hat\widehat

\renewcommand{\arraystretch}{1.15}


%% file: 1_title.tex
Linear-time Online Visual Geometry Transformer with Hybrid Architecture

%% file: 2_authors.tex
\begin{icmlauthorlist}
\mbox{Haoyu Kang$^{\,1\,,\textrm{\Letter}}$},
\mbox{Pingzhi Li$^{\,1\,}$}, 
\mbox{Ruichen Zhang$^{\,1\,}$}, 
and \mbox{Tianlong Chen$^{\,1\,,\textrm{\Letter}}$}
\end{icmlauthorlist}

$^{1\,}$UNITES Lab, University of North Carolina at Chapel Hill 

\quad $^{\textrm{\Letter}}$ Corresponding Author

%% file: tex/0_abs.tex
Transformer-based 3D reconstruction models such as VGGT and $\pi^3$ predict cameras, depth and point maps directly from many input views. Causal attention models such as StreamVGGT extend them to online video but cache all past tokens, so their total compute grows quadratically with the stream length. Recurrent models such as CUT3R and TTT3R keep a fixed-size state at constant cost, but they progressively forget early geometry. We present LinearVGGT, a hybrid model distilled from StreamVGGT, which uses Mamba2 layers for efficient sequence modeling in three of every four global layers. The remaining layers use register-guided sparse attention to retrieve a few relevant past frames, compensating for the limited exact long-range recall of Mamba2. Our distillation needs only 9\% of the GPU hours and 31\% of the training frames that StreamVGGT uses to learn from VGGT. LinearVGGT keeps an almost constant latency of 27 ms per frame over 1,024 frames and is on average $12 \times$ faster than StreamVGGT. It outperforms its teacher and is competitive with the state-of-the-art streaming model LingBot-Map, while running $2.5\times$ faster and using $2.7\times$ less memory.

%% file: tex/1_intro.tex
\section{Introduction}

\label{sec:intro}

Recent advances in transformer-based 3D visual geometry reconstruction have demonstrated strong performance in recovering scenes’ 3D  geometry attributes (e.g., camera parameters, point maps, depth maps) from multiple views of  scenes~\cite{wang2025vggt,wang2025pi3,wang2026vggtomega}.
However, their reliance on dense global attention incurs quadratic computational complexity in the number of visual tokens and substantial memory overhead, limiting their scalability to long image sequences and applications to streaming video inputs like SLAM~\cite{maggio2025vggt-slam}.

These limitations have motivated a growing body of work aimed at improving the computational and memory efficiency of these models. Token-merging based methods like FastVGGT,  GoToHunt etc recognize the redundant attention pattern in global attention and only select the most important tokens for computation. Despite being easy to implement and training free, these methods still struggle with quadratic computational complexity of global attention. Recurrent models such as CUT3R~\cite{wang2025cut3r} and TTT3R~\cite{chen2025ttt3r} keep a fixed-size state at a constant cost per frame, but this state forgets earlier frames as the stream grows~\cite{chen2025ttt3r}. Furthermore, StreamVGGT~\cite{zhuo2025streamvggt} and LingBot-Map~\cite{chen2026lingbotmap} have explored using causal attention architectures or its more efficient variants to solve this. Nevertheless, the nature of causal attention determines that it still suffers from prohibitively long inference time when increasing the number of processed frames. Attention-based models thus stay accurate, but their cost per frame grows with the stream.
Fixed-size states run at a constant cost, but they cannot keep every past frame.
This tradeoff raises our research question: \textbf{\textit{Can a streaming visual geometry transformer run at a nearly constant cost per frame without losing the accuracy of causal attention?}}

\begin{figure}
    \centering
    \includegraphics[width=\linewidth]{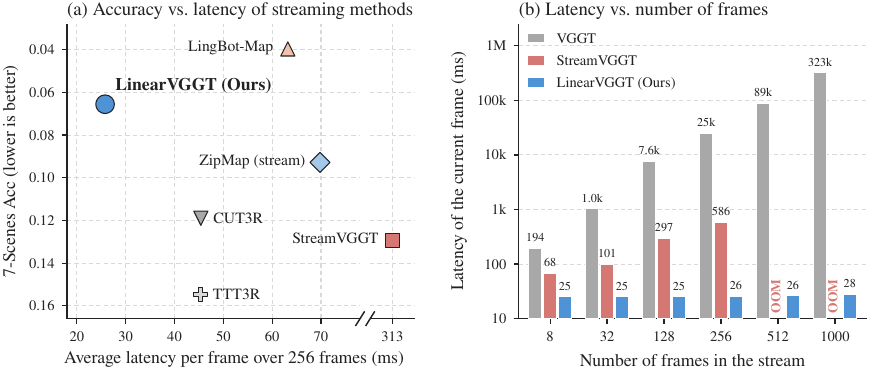}
    \caption{(a) 7-Scenes accuracy against the average latency per frame over a 256-frame stream for streaming methods.  (b) Latency of the \textbf{current frame} as the stream grows; VGGT processes all frames again for each new frame.}
    \label{fig:efficiency_overview}
    \vspace{-1.5em}
\end{figure}

Linear-time sequence models such as Mamba2~\cite{dao2024mamba2} and Gated DeltaNet~\cite{yang2025gdn} suggest an answer. They carry the history in a fixed-size state, so each new input costs the same, and hybrid language models~\cite{lieber2024jamba,kimi2025linear} recover the recall of attention with a few attention layers among them. Although the great efficiency makes it appealing to adapt these cutting-edge architectures, applying them to 3D reconstruction domain is non-trivial, we recognize two primary challenges:
First, the adaptation to 3D reconstruction domains, especially performance on long sequences. We observe significant performance drop of selective state space models under long sequences inputs, which counteracts the efficiency gain of these linear models.
Second, the new modeling architecture will trigger training from scratch, which requires extensive GPU time and resources (~\cref{tab:cost}.)

To address these challenges,  we propose LinearVGGT,  a hybrid streaming visual geometry transformer whose Mamba2 layers carry a compressed summary of the stream at a constant cost. To counteract the decay of this summary on long streams, we interleave the Mamba2 layers with periodic attention layers in a 3:1 ratio. These attention layers fill in what the summary misses: direct interactions with the key tokens of the past frames most relevant to the new frame.
Full attention in these layers would bring the growing cost back, yet most of its interactions are redundant.
Prior analyses of VGGT show that its global attention can be sparsified  with little loss of accuracy~\cite{wang2026blocksparse, sun2026avggt}.
Our preliminary experiments agree: a new frame loses little accuracy when it reads only the first frame and a few past frames (\cref{sec:analysis_exp}).
The open question is which frames and tokens to read.
Register tokens offer a cheap answer: in vision transformers they collect global information about their image~\cite{darcet2024registers}, and VGGT-$\Omega$ already uses them to exchange scene information across frames~\cite{wang2026vggtomega}.
We therefore propose register-guided sparse attention, a hierarchical design that selects frames first and tokens second.
At the frame level, the registers of a new frame rank all past frames by attending only to their registers, and the new frame attends to the first frame, which anchors the world coordinates, and to the top-ranked frames.
At the token level, a lightweight indexer keeps in each past frame only the patch tokens that the frame itself attends to most.

To avoid the training cost discussed above, we distill LinearVGGT from StreamVGGT with a three-stage procedure inspired by MOHAWK~\cite{bick2024mohawk}.
Starting from the attention weights of the teacher, the first two stages align each new layer with the corresponding attention layer, first through its attention matrix and then through its output.
The third stage trains the student end to end against both the predictions of the teacher and ground-truth depth, cameras and point maps, first updating the new layers and then all layers except the image encoder. The whole procedure takes 59 hours on one GPU and 1.2M training frames.
This is 9\% of the GPU hours and 31\% of the frames that StreamVGGT needs to learn from VGGT~\cite{zhuo2025streamvggt}, and less than 0.5\% of the GPU hours used to train VGGT or LingBot-Map.
Despite this small budget, LinearVGGT has lower accuracy and completeness errors than its teacher in all four multi-view reconstruction settings, with up to 49\% lower accuracy error.
It also outperforms the recurrent CUT3R and TTT3R. Additionally, we observe that streaming inference is also limited by the overhead of launching GPU kernels. We therefore optimize inference by capturing the fixed-shape part of each frame as CUDA graphs and fusing its small kernels. With these optimizations, LinearVGGT keeps its latency of average 27ms per frame, on average 12 times faster than StreamVGGT and also faster than CUT3R and TTT3R.

Our contributions can be summarized as follows:

\begin{itemize}
  \item We propose LinearVGGT, a streaming visual geometry transformer that interleaves Mamba2 layers with register-guided sparse attention, so that its cost stays nearly constant while it keeps the accuracy of causal attention;
  \item We introduce a three-stage distillation for visual geometry transformers that turns a pretrained model into an accurate linear-time model with a small fraction of the data and compute that training such a model usually requires;
  \item Extensive experiments show that LinearVGGT reaches state-of-the-art efficiency in streaming reconstruction while its accuracy remains competitive with that of state-of-the-art models.
\end{itemize}

%% file: tex/3_method.tex
\section{Method}

\begin{figure}[!t]
    \centering
    \includegraphics[width=\linewidth]{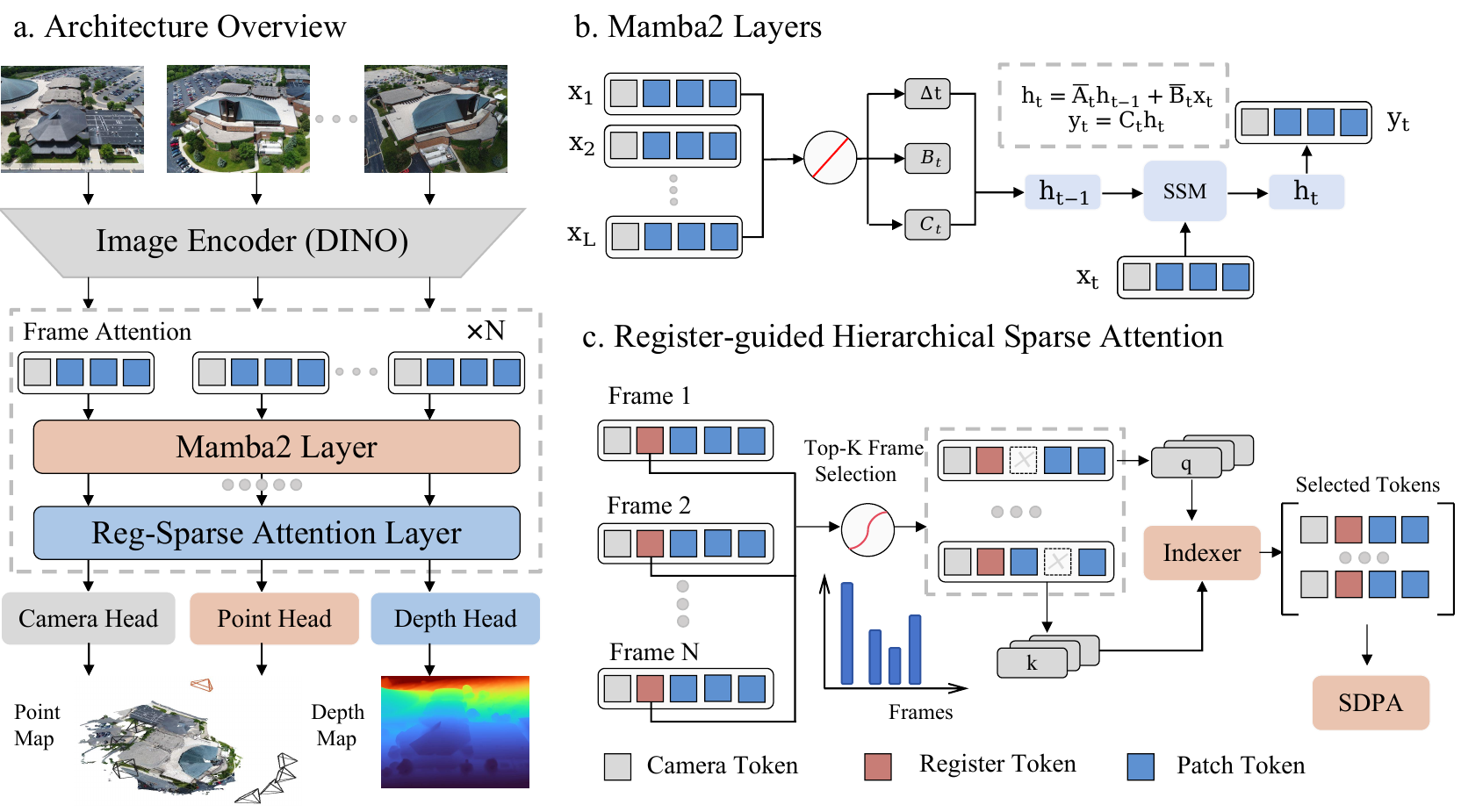}
    \caption{Overview of the architecture of LinearVGGT: a. LinearVGGT interleaves Mamba2 layers with register-guided sparse attention layers. b. A Mamba2 layer carries the history in a fixed-size state $h_t$, which it updates with the tokens of each new frame. c. Details of register-guided sparse attention layers}
    \label{fig:overview}
\end{figure}

\Cref{fig:overview} gives an overview of LinearVGGT.
It keeps the image encoder, the frame attention layers and the prediction heads of StreamVGGT and changes only the layers that connect frames: three of every four become Mamba2 layers (\cref{sec:mamba}), and every fourth becomes a register-guided sparse attention layer (\cref{sec:sparse}).
We obtain the model by distillation from StreamVGGT (\cref{sec:distill}) and accelerate its streaming inference (\cref{sec:inference}).

\subsection{Preliminaries: Streaming 3D Reconstruction}
\label{sec:prelim}

\paragraph{Problem setting.}
A streaming model receives RGB frames $I_0, I_1, \dots$ one at a time, without camera parameters.
For each frame $t$, it predicts the camera $\mathbf{g}_t$, the depth map $D_t$ and the point map $X_t$ in the coordinate frame of $I_0$, from the frames $I_0, \dots, I_t$ alone and before frame $t{+}1$ arrives.
The model must therefore carry what it needs from the past in some memory, and the cost of reading this memory decides how long a stream it can process.

\paragraph{Causal Global Attention}
StreamVGGT~\cite{zhuo2025streamvggt}, the model we build on, keeps the whole past as this memory.
Following VGGT~\cite{wang2025vggt},
it encodes each frame into $P$ patch tokens and adds one camera token and $R$ register tokens, which gives $N = P + R + 1$ tokens per frame.
An aggregator of $L$ blocks then alternates frame attention, among the tokens of one frame, with global attention, across frames, and two heads decode the camera token and the patch tokens.
In streaming mode, the global attention is causal over frames and caches the keys and values of every past frame.
For one head of a global layer, the tokens of frame $t$ attend to all cached frames:
\begin{equation}
  O_t = \operatorname{softmax}\!\Big(\tfrac{1}{\sqrt{d}}\,Q_t\,[K_0; \dots; K_t]^\top\Big)\,[V_0; \dots; V_t],
  \label{eq:global}
\end{equation}
where $Q_t, K_s, V_s \in \mathbb{R}^{N \times d}$ are the queries of frame $t$ and the keys and values of frame $s$.
Frame $t$ therefore costs $\mathcal{O}(tN^2 d)$ in every global layer, and the cache of every layer grows by $N$ tokens per frame.

\subsection{Mamba2 layers for efficient stream modeling}
\label{sec:mamba}

To remove this growth, most global layers of LinearVGGT carry the past in a state of fixed size.
In these layers, we replace the attention operator with a Mamba2 mixer~\cite{dao2024mamba2} and keep the normalization, residual connection and MLP of the block.
The mixer reads the tokens of all frames as one sequence $u_1, u_2, \dots$ in temporal order: frame by frame, and inside a frame the camera token, the register tokens and then the patch tokens in raster order.
For each head, it updates a state $H_i \in \mathbb{R}^{d \times n}$ as
\begin{equation}
  H_i = e^{\Delta_i a}\, H_{i-1} + \Delta_i\, \mathbf{x}_i \mathbf{b}_i^\top, \qquad \mathbf{y}_i = H_i\, \mathbf{c}_i,
  \label{eq:ssm}
\end{equation}
where every quantity with index $i$ is computed from the token $u_i$.
The token writes its content $\mathbf{x}_i \in \mathbb{R}^{d}$ into the state under the key $\mathbf{b}_i \in \mathbb{R}^{n}$ and reads the output $\mathbf{y}_i \in \mathbb{R}^{d}$ from it with the query $\mathbf{c}_i \in \mathbb{R}^{n}$, so that $\mathbf{x}$, $\mathbf{b}$ and $\mathbf{c}$ play the roles of the value, key and query of attention.
The step $\Delta_i > 0$ sets both how strongly the token is written and, with the decay rate $a < 0$ learned per head, how much of the earlier state is kept.
Unrolling \cref{eq:ssm} gives
\begin{equation}
  \mathbf{y}_i = \sum_{j \le i} M_{ij}\, \mathbf{x}_j, \qquad M_{ij} = \mathbf{c}_i^\top \mathbf{b}_j \, \exp\!\Big(a \!\sum_{m=j+1}^{i}\! \Delta_m\Big)\, \Delta_j ,
  \label{eq:mixing}
\end{equation}
so the mixer acts as a causal linear attention whose weights decay with the distance between two tokens, at a rate that each token sets through $\Delta$.
During streaming, only the state $H$ carries over from one frame to the next.
A frame thus costs $\mathcal{O}(N)$ per layer, and the memory of a Mamba2 layer does not depend on $t$.
The same decay, however, is what makes the state forget: once the weight $M_{ij}$ of a distant token has vanished, no later token can read it again. A stack of Mamba2 layers alone would thus lose the early frames of a long stream, which is the limitation that the next section addresses.

\subsection{Register-guided hierarchical sparse attention}
\label{sec:sparse}

A geometry model needs more from the past than a decaying summary.
The first frame defines the world coordinates of every prediction, and a frame that revisits a place must align with the earlier views of it.
We therefore follow hybrid language models~\cite{lieber2024jamba,kimi2025linear} and interleave the Mamba2 layers with attention layers in a 3:1 ratio, which keep the weights of the corresponding teacher layers.
Full attention in these layers would bring back the cost of \cref{eq:global}, but most of it is redundant~\cite{wang2026blocksparse, sun2026avggt}, so we sparsify it with a two-level hierarchy: a new frame first selects a few past frames and then attends only to a subset of their tokens.

\textbf{Frame routing with register tokens.}
Scoring every past frame with all $N$ tokens of the new frame would itself cost as much as full attention.
Register tokens offer a cheaper signal: they collect global information about their image~\cite{darcet2024registers}, so the attention between the registers of two frames indicates how much one frame needs the other.
Let $\mathbf{q}^h_{t,i}$ be the query of register $i$ of frame $t$ in head $h$, and $\mathbf{k}^h_{s,j}$ the key of register $j$ of frame $s$.
The routing score of frame $s$ is the attention mass that the registers of frame $t$ place on the registers of frame $s$, averaged over the $H$ heads and the $R$ register queries:
\begin{equation}
  \pi_{t,s} = \frac{1}{HR} \sum_{h=1}^{H} \sum_{i=1}^{R} \sum_{j=1}^{R} \operatorname{softmax}_{(s',j')} \Big(\tfrac{1}{\sqrt{d}}\, \mathbf{q}^{h\top}_{t,i}\, \mathbf{k}^h_{s',j'}\Big)_{(s,j)},
  \label{eq:route}
\end{equation}
where the softmax runs over the register keys of frames $0, \dots, t$.
Frame $t$ then attends to the set
\begin{equation}
  \mathcal{S}_t = \{0,\, t\} \,\cup\, \operatorname{top-}\!k_{\,1 \le s < t}\; \pi_{t,s},
  \label{eq:select}
\end{equation}
which contains the first frame as a fixed anchor, the frame itself and the $k$ past frames with the highest scores.
The routing reuses the queries and keys that the attention layer computes anyway, so it adds no parameters.

\textbf{Token indexing.}
A selected frame still brings $N$ tokens, many of which cover textureless or repeated regions.
When frame $s \ge 1$ enters the cache of a sparse layer, a token indexer therefore keeps its camera and register tokens and only the $\lceil \rho P \rceil$ patch tokens that receive the most attention from the tokens of frame $s$ itself, summed over heads and queries.
This score is also a by-product of the layer and needs no parameters. The anchor frame keeps all its tokens, and the register keys of every frame stay in the cache.

\textbf{Cost.}
With this hierarchy design, frame $t$ attends to $2N + k(R{+}1{+}\lceil\rho P\rceil)$ keys in each sparse layer for any $t$.
Only the $R$ register queries read the whole history, at $R$ keys per past frame, about $N^2/R^2$ times fewer query-key pairs than \cref{eq:global}.
In contrast to VGGT-$\Omega$~\cite{wang2026vggtomega}, whose register attention lets only registers exchange information across frames, our registers only select the frames, and all tokens of the new frame then attend to the selected tokens.

\subsection{Three-Stage Distillation}
\label{sec:distill}

Training a new architecture from scratch requires the compute and data of a large-scale pretraining run, most of which would go into relearning what StreamVGGT already knows.
Since only the cross-frame layers of LinearVGGT are new, we keep the pretrained weights and teach these layers to behave like the attention they replace.
Inspired by MOHAWK~\cite{bick2024mohawk}, we transfer the teacher into the student in three stages that move from the operator to the output.
\Cref{eq:mixing} provides the starting point: we initialize the maps that produce $\mathbf{x}$, $\mathbf{b}$ and $\mathbf{c}$ with the value, key and query projections of the teacher layer and copy its output projection, and the sparse layers start as exact copies of their teacher layers.

\textbf{Stage 1: attention matrices.}
With the inputs of every layer taken from the teacher, each Mamba2 layer learns to reproduce the attention matrix $A$ of its teacher layer with its mixing matrix $M$ of \cref{eq:mixing}, on a set $\Omega$ of sampled query rows:
\begin{equation}
  \mathcal{L}_1 = \| M_\Omega - A_\Omega \|_F^2 \,/\, \| A_\Omega \|_F^2 ,
\end{equation}
averaged over the Mamba2 layers.

\textbf{Stage 2: layer outputs.}
Still with teacher inputs, each Mamba2 layer matches the output $\mathbf{o}$ of the teacher attention through the relative error $\mathcal{L}_2 = \|\mathbf{y} - \mathbf{o}\|^2 / \|\mathbf{o}\|^2$.
After these two stages, every new layer approximates its teacher layer in isolation.

\textbf{Stage 3: end to end.}
The errors of the individual layers accumulate when all of them run together, so the last stage trains the student on its predictions:
\begin{equation}
  \mathcal{L}_3 = \mathcal{L}_{\text{GT}} + \lambda\, \mathcal{L}_{\text{KD}} .
  \label{eq:stage3}
\end{equation}
$\mathcal{L}_{\text{KD}}$ penalizes the difference between the depth maps, point maps and cameras of the student and those of the teacher on the same clip, and $\mathcal{L}_{\text{GT}}$ compares the same outputs with scale-normalized ground truth.
A student trained only to imitate inherits the errors of its teacher, and the ground-truth term lets it correct them.
We first update the new layers alone and then all layers except the image encoder, so that the frame attention layers and the heads adapt to the new cross-frame layers.

\subsection{Efficient streaming inference}
\label{sec:inference}

Streaming inference is bound by kernel launches rather than by computation (\cref{sec:analysis}), and a CUDA graph removes this overhead by replaying a recorded sequence of kernels with a single launch, provided that every tensor keeps its shape.
In LinearVGGT, the fixed-size states of the Mamba2 layers make this hold for almost the whole frame, so we capture it as seven CUDA graphs, split at the six sparse attention layers, and run only these layers and the camera head, whose inputs grow with the stream, kernel by kernel.
We also cache the positional embeddings and the normalized, rotated keys of past frames, and fuse sequences of small operations into single kernels.

%% file: tex/4_experiments.tex
\section{Experiment}

Our experiments answer three questions: \textbf{Q1:} How accurate is LinearVGGT compared with state-of-the-art streaming models, including its attention-based teacher? \textbf{Q2:} Does its cost per frame stay constant as the stream grows, and what does it cost to train? \textbf{Q3:} Which parts of the design are responsible?

\input{table/3d_reconstruction}

\subsection{Setup}

\paragraph{Training.}
We train on 14 datasets that span indoor, outdoor, object-centric and synthetic scenes, including ScanNet~\cite{dai2017scannet}, CO3Dv2~\cite{reizenstein2021co3d}, MegaDepth~\cite{li2018megadepth} and Waymo~\cite{sun2020waymo}.
The supplementary material lists all datasets, the hardware and the remaining training settings.

\textbf{Baselines.}
We compare with the streaming models StreamVGGT~\cite{zhuo2025streamvggt}, OVGGT~\cite{lu2026ovggt}, CUT3R~\cite{wang2025cut3r}, TTT3R~\cite{chen2025ttt3r}, LingBot-Map~\cite{chen2026lingbotmap} and ZipMap~\cite{jin2026zipmap} in its streaming mode.
As a reference, we also list offline models that see all frames at once: VGGT~\cite{wang2025vggt}, $\pi^3$~\cite{wang2025pi3}, FastVGGT~\cite{shen2025fastvggt} and GoToHunt~\cite{zheng2026gotohunt}.
We run every baseline with its official code and weights in the same evaluation pipeline.

\textbf{Benchmarks and metrics.}
For multi-view reconstruction, we follow the protocol of CUT3R and StreamVGGT on 7-Scenes~\cite{shotton2013sevenscenes} and NRGBD~\cite{azinovic2022neuralrgbd} and report accuracy (Acc~$\downarrow$), completeness (Comp~$\downarrow$) and normal consistency (NC~$\uparrow$).
With \emph{sparse views}, we take one keyframe every 200 frames on 7-Scenes and every 500 frames on NRGBD, which gives 2 to 5 frames per sequence; with \emph{dense views}, one every 20 and 50 frames, which gives 8 to 50 frames.
For video depth, we follow the protocol of CUT3R on Sintel~\cite{butler2012sintel}, Bonn~\cite{palazzolo2019refusion} and KITTI~\cite{geiger2013kitti}, which cover synthetic, indoor and driving scenes with moving objects.
We align the predicted depth of each sequence to the ground truth, either with one scale factor or with a scale and a shift, and report the absolute relative error (Abs~Rel~$\downarrow$) and the share of pixels with $\delta < 1.25$ ($\uparrow$).

\subsection{3D Reconstruction}
\label{sec:recon}

We first evaluate scene-level reconstruction from sparse views, which tests each model on image collections with little overlap.
As \cref{tab:recon} shows, among the online methods, LinearVGGT outperforms its teacher StreamVGGT and the recurrent CUT3R and TTT3R in accuracy and completeness on both datasets, and it halves the accuracy error of StreamVGGT on 7-Scenes.
Only LingBot-Map is more accurate, while LinearVGGT runs 2.5 times faster than it and 12 times faster than StreamVGGT.
Compared with the offline methods, which see all frames at once, LinearVGGT is more accurate on 7-Scenes and on par with VGGT on NRGBD, at a higher throughput than any of them.
With sparse inputs, LinearVGGT thus combines an accuracy competitive with state-of-the-art online and offline models with the highest throughput in the comparison.

\begin{figure}[h]
    \centering
    \includegraphics[width=\columnwidth]{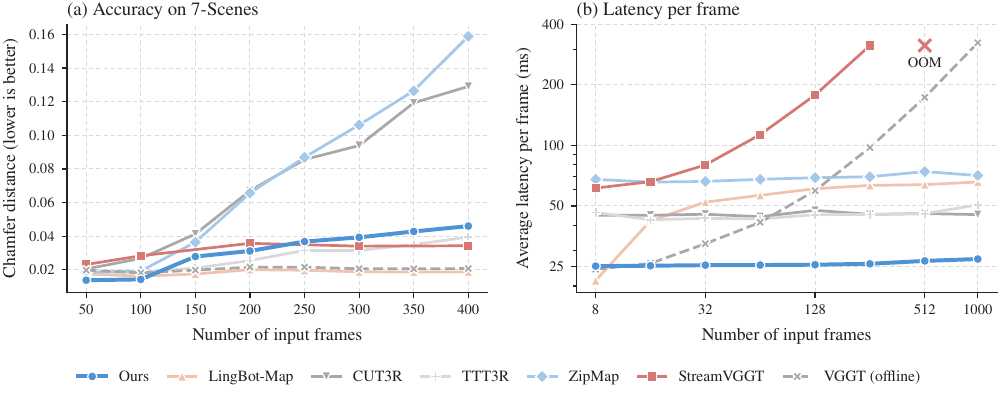}
    \caption{Accuracy and efficiency vs. sequence length. The Chamfer distance is the mean of accuracy and completeness. }
    \label{fig:acc_vs_frames}
\end{figure}

Sparse views show how a model handles little overlap, but not how it behaves as a stream grows.
We therefore also evaluate dense streams: we take every second frame of each sequence and reconstruct the first $N$ of them, for increasing $N$ (\cref{fig:acc_vs_frames}). The error of LinearVGGT also grows with the length of the stream, from 0.014 at 50 frames to 0.046 at 400, but far more slowly than that of CUT3R and streaming ZipMap, which reach 0.129 and 0.159, and at a rate similar to TTT3R and StreamVGGT. Only LingBot-Map stays flat over the whole range. LinearVGGT models long sequences less well than it, but is the fastest method in the comparison: over 256 frames it runs 2.5 times faster than LingBot-Map, 1.8 times faster than CUT3R and TTT3R, and 12 times faster than StreamVGGT.

\begin{figure}[tbp]
    \centering
    \includegraphics[width=\linewidth]{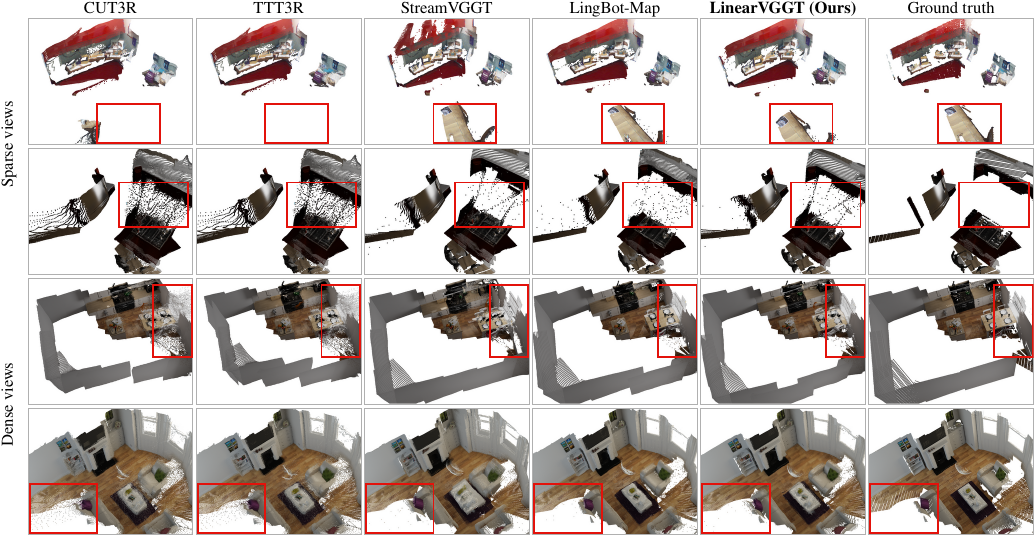}
    \caption{Qualitative Results for 3D Reconstruction. }
    \label{fig:qualitative}
\end{figure}

\Cref{fig:qualitative} shows reconstructions with sparse and dense views.
With sparse views, CUT3R and TTT3R misplace or lose a part of the scene that only one distant view observes, whereas LinearVGGT places it as in the ground truth.
With dense views, they leave stray points along walls and floors, which LinearVGGT largely avoids; its reconstructions are close to those of LingBot-Map.

\subsection{Video Depth Estimation}
\input{table/video_depth}
We next evaluate video depth, which tests both the quality of the depth of each frame and its consistency over a video with moving objects.
As \cref{tab:vdepth} shows, LinearVGGT again outperforms its teacher: it lowers the absolute relative error of StreamVGGT on KITTI from 0.173 to 0.093 and on Bonn from 0.059 to 0.052, and matches it on Sintel.
It is also more accurate than the recurrent CUT3R and TTT3R on all three datasets, by a wide margin on Sintel and KITTI.
Among the online methods, LinearVGGT achieves the best or second-best absolute relative error on all three datasets, ranking first on Bonn and second on Sintel and KITTI.
On Bonn, its error is also lower than that of the offline VGGT.
The gains of LinearVGGT are thus not limited to static indoor scenes: its per-frame geometry stays accurate and consistent on dynamic and outdoor videos as well.

\subsection{Analysis}
\label{sec:analysis}

\input{table/cost}

\textbf{Inference efficiency.}
\label{sec:efficiency}
The latency of LinearVGGT stays between 25 and 28\,ms per frame over a stream of 1,024 frames (\cref{fig:efficiency_overview}b).
The latency of StreamVGGT grows from 68\,ms at frame 8 to 586\,ms at frame 256, and the model runs out of memory after 445 frames on a 96\,GB GPU.
The recurrent CUT3R and TTT3R also run at a constant latency, of about 45\,ms, and LingBot-Map takes 59 to 71\,ms, so LinearVGGT is the fastest of the online models at every stream length.
Its memory grows slowly: the caches of the sparse attention layers hold 6.3\,GiB after 1,024 frames, about 6\,MiB per frame, against 98\,MiB per frame for the cache of StreamVGGT.

\begin{figure}[t]
\centering
\includegraphics[width=0.6\linewidth]{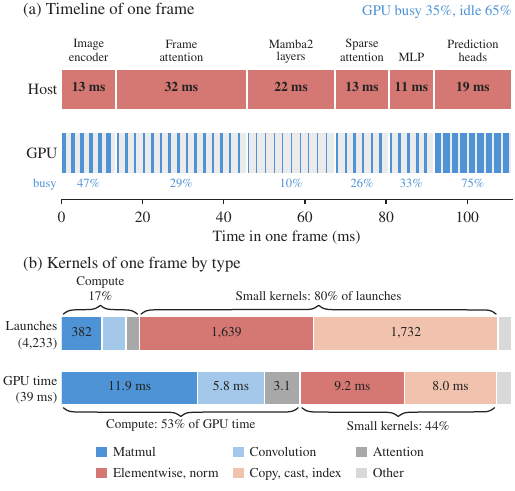}
\caption{\textbf{Streaming inference is limited by kernel launches.} Time that the host spends on launching GPU kernels and GPU computation time, for one frame of LinearVGGT.}
\label{fig:launch}
\end{figure}

\textbf{Inference optimization.}
\Cref{fig:launch} shows where the time of a frame goes in eager execution.
In every module except the prediction heads, the host spends several times longer on launching GPU kernels than the GPU spends on computing them: a frame launches about 4,200 kernels, and the GPU is busy for only 35\% of the time.
The gap is largest in the Mamba2 layers, whose computation takes 2\,ms of the 22\,ms they occupy.
Launches dominate because a fixed-size state keeps the work of each frame small and constant.
In StreamVGGT, attention over the growing cache takes over instead: the same engine lowers its latency from 62 to 40\,ms at frame 8, but only from 584 to 574\,ms at frame 256.
We therefore capture the fixed-shape work of each frame as CUDA graphs, cache the repeated computation and fuse small kernels (\cref{sec:inference}).
These optimizations lower the latency of a frame from 78 to 27\,ms.

\textbf{Training efficiency.}
\Cref{tab:cost} compares the training cost of LinearVGGT with that of related models.
Our distillation takes 59 GPU hours and 1.2M training frames, which is 9\% of the GPU hours and 31\% of the frames that StreamVGGT needs to learn from VGGT.
Models with a new architecture that are trained from scratch, such as VGGT and LingBot-Map, need more than 200 times as many GPU hours.

\begin{figure}[t]
\centering
\includegraphics[width=0.6\linewidth]{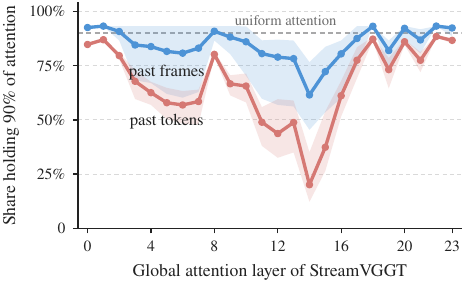}
\caption{\textbf{Global attention is redundant.} Fraction of past frames and of past tokens that a new frame needs to cover 90\% of its attention, in each global layer of StreamVGGT.}
\label{fig:redundancy}
\end{figure}

\textbf{Redundancy of attention.}
\label{sec:analysis_exp}
Register-guided sparse attention rests on the premise that a new frame does not need to attend to the whole history.
\Cref{fig:redundancy} examines this premise on the global attention of StreamVGGT, for clips of 16 frames.
In the first and last layers, the attention of a new frame is close to uniform over the past, so it averages the history rather than retrieves from it.
In the middle layers, the attention concentrates: in layer 14, a fifth of the past tokens holds 90\% of it.
In neither case does a new frame need all tokens of all past frames.
The trained model agrees.
Letting its sparse layers read all past frames with all their tokens does not improve accuracy: with dense views, the mean error stays level on NRGBD (0.045 against 0.048) and rises on 7-Scenes from 0.037 to 0.062.
It only raises the cost: after 1,024 frames, a frame takes 335\,ms instead of 25\, - 28\,ms.
A few selected frames thus suffice, at a fraction of the cost of full attention.

\subsection{Ablation Study}

\input{table/ablation_recon}

We remove or replace one component of LinearVGGT at a time and report the accuracy with dense views (\cref{tab:ablation}).
We use dense views because, with sparse views, the anchor and the selected frames already cover the whole input.

\textbf{Hybrid layout.}
A student with Mamba2 layers in all global layers has an accuracy error five times as high on 7-Scenes (0.185 against 0.037) and four times as high on NRGBD (0.197 against 0.048) as a student that keeps attention in every fourth layer. A recurrent state alone thus does not carry what the reconstruction needs from the past, which confirms the need for the sparse attention layers.

\textbf{Anchor and frame routing.}
Without the anchor, the accuracy error rises from 0.037 to 0.049 on 7-Scenes and from 0.048 to 0.055 on NRGBD, although the model reads the same number of frames.
The first frame defines the coordinate frame of all predictions, and the routing alone does not select it reliably.
 For the remaining frames, register attention beats both alternatives: the most recent frames raise the error to 0.046 and 0.054, and random frames to 0.079 and 0.050. Which past frames are read thus matters.

\textbf{Token indexer.}
The purpose of the indexer is to save memory.
Without it, every cached frame keeps all its tokens, and the caches of the sparse layers hold 24.5\,GiB after 1,024 frames; the indexer lowers this to 6.3\,GiB.
Accuracy and completeness stay at the same level, so the tokens it removes are not needed.
Keeping the tokens that a frame attends to most is more accurate than keeping random or evenly spaced tokens on 7-Scenes and equal on NRGBD.

\textbf{Ground-truth supervision.}
Removing the ground-truth term from Stage 3 increases the mean accuracy error from 0.037 to 0.106 on 7-Scenes. These results demonstrate the importance of combining teacher distillation with direct geometric supervision in our training setup.

%% file: table/3d_reconstruction.tex
\begin{table}[!t]
\centering
\caption{\textbf{3D reconstruction with sparse views on 7-Scenes and NRGBD.} FPS is the average over a stream of 256 frames. Among online methods, the best result is in \textbf{bold} and the second best is \underline{underlined}.}
\label{tab:recon}
\footnotesize
\setlength{\tabcolsep}{2.8pt}
\renewcommand{\arraystretch}{0.92}
\resizebox{\linewidth}{!}{%
\begin{tabular}{l c cc cc cc cc cc cc c}
\toprule
\multirow{3}{*}[-0.6ex]{\textbf{Method}} & \multirow{3}{*}[-0.6ex]{\textbf{Mode}} & \multicolumn{6}{c}{\textbf{7-Scenes}~\cite{shotton2013sevenscenes}} & \multicolumn{6}{c}{\textbf{NRGBD}~\cite{azinovic2022neuralrgbd}} & \multirow{3}{*}[-0.6ex]{\textbf{FPS}} \\
\cmidrule(lr){3-8} \cmidrule(lr){9-14}
 & & \multicolumn{2}{c}{Acc$\downarrow$} & \multicolumn{2}{c}{Comp$\downarrow$} & \multicolumn{2}{c}{NC$\uparrow$} & \multicolumn{2}{c}{Acc$\downarrow$} & \multicolumn{2}{c}{Comp$\downarrow$} & \multicolumn{2}{c}{NC$\uparrow$} & \\
\cmidrule(lr){3-4} \cmidrule(lr){5-6} \cmidrule(lr){7-8} \cmidrule(lr){9-10} \cmidrule(lr){11-12} \cmidrule(lr){13-14}
 & & Mean & Med. & Mean & Med. & Mean & Med. & Mean & Med. & Mean & Med. & Mean & Med. & \\
\midrule
VGGT~\cite{wang2025vggt}                   & Offline & 0.088             & 0.039             & 0.091             & 0.040             & 0.786             & 0.889             & 0.079             & 0.021             & 0.083             & 0.024             & 0.908             & 0.989             & 10.3 \\
$\pi^3$~\cite{wang2025pi3}                 & Offline & 0.123             & 0.036             & 0.115             & 0.037             & 0.783             & 0.888             & 0.028             & 0.013             & 0.030             & 0.012             & 0.910             & 0.992             & 14.1 \\
\midrule
FastVGGT~\cite{shen2025fastvggt}           & Offline & 0.091             & 0.043             & 0.087             & 0.034             & 0.738             & 0.847             & 0.101             & 0.038             & 0.100             & 0.041             & 0.859             & 0.959             & 23.7 \\
GoToHunt~\cite{zheng2026gotohunt}          & Offline & 0.075             & 0.036             & 0.070             & 0.034             & 0.757             & 0.873             & 0.060             & 0.022             & 0.069             & 0.029             & 0.893             & 0.990             & 20.4 \\
\midrule
StreamVGGT~\cite{zhuo2025streamvggt}       & Online  & 0.129             & 0.055             & 0.115             & 0.040             & 0.751 & \underline{0.867} & 0.084             & 0.044             & 0.079             & 0.038             & 0.861 & \underline{0.986} & 3.2 \\
OVGGT~\cite{lu2026ovggt}                   & Online  & 0.128             & 0.054             & 0.114             & 0.040             & \underline{0.752}             & \underline{0.867}             & 0.084             & 0.044             & 0.079             & 0.038             & \underline{0.862}             & \underline{0.986}             & 10.3 \\
CUT3R~\cite{wang2025cut3r}                 & Online  & 0.119             & 0.049             & 0.120             & \underline{0.031} & 0.722             & 0.826             & 0.094             & \underline{0.027} & 0.073             & \underline{0.024} & 0.836             & 0.970             & 22.0 \\
TTT3R~\cite{chen2025ttt3r}                 & Online  & 0.155             & 0.049             & 0.194             & 0.043             & 0.715             & 0.815             & 0.101             & 0.032             & 0.076             & 0.027             & 0.838             & 0.970             & \underline{22.1} \\
ZipMap~\cite{jin2026zipmap}                & Online  & 0.093             & 0.064             & 0.107             & 0.043             & 0.742             & 0.850             & 0.217             & 0.167             & 0.212             & 0.171             & 0.809             & 0.911             & 14.3 \\
LingBot-Map~\cite{chen2026lingbotmap}      & Online  & \textbf{0.040}    & \textbf{0.022}    & \textbf{0.061}    & \textbf{0.023}    & \textbf{0.768}    & \textbf{0.882}    & \textbf{0.039}    & \textbf{0.018}    & \textbf{0.045}    & \textbf{0.016}    & \textbf{0.893}    & \textbf{0.987}    & 15.8 \\
\textbf{LinearVGGT (Ours)}                 & Online  & \underline{0.066} & \underline{0.032} & \underline{0.073} & \underline{0.031} & 0.748             & 0.858             & \underline{0.078} & \underline{0.027} & \underline{0.067} & \underline{0.024} & 0.859             & 0.985             & \textbf{38.8} \\
\bottomrule
\end{tabular}}
\end{table}

%% file: table/video_depth.tex
\begin{table}[t]
\centering
\caption{\textbf{Video depth estimation.} The predicted depth of each sequence is aligned to the ground truth with a scale factor. Among online methods, the best result is in \textbf{bold} and the second best is \underline{underlined}.}
\label{tab:vdepth}
\footnotesize
\setlength{\tabcolsep}{5pt}
\renewcommand{\arraystretch}{0.92}
\begin{tabular}{l c cc cc cc}
\toprule
 & & \multicolumn{2}{c}{\textbf{Sintel}} & \multicolumn{2}{c}{\textbf{Bonn}} & \multicolumn{2}{c}{\textbf{KITTI}} \\
\cmidrule(lr){3-4} \cmidrule(lr){5-6} \cmidrule(lr){7-8}
\textbf{Method} & \textbf{Mode} & AbsRel$\downarrow$ & $\delta_{1.25}\!\uparrow$ & AbsRel$\downarrow$ & $\delta_{1.25}\!\uparrow$ & AbsRel$\downarrow$ & $\delta_{1.25}\!\uparrow$ \\
\midrule
VGGT~\cite{wang2025vggt}                 & Offline & 0.298             & 0.682             & 0.057             & 0.968             & 0.061             & 0.970 \\
$\pi^3$~\cite{wang2025pi3}               & Offline & 0.248             & 0.679             & 0.049             & 0.975             & 0.037             & 0.986 \\
\midrule
FastVGGT~\cite{shen2025fastvggt}         & Offline & 0.297             & 0.675             & 0.056             & 0.971             & 0.064             & 0.968 \\
GoToHunt~\cite{zheng2026gotohunt}        & Offline & 0.294             & 0.671             & 0.060             & 0.965             & 0.063             & 0.971 \\
\midrule
StreamVGGT~\cite{zhuo2025streamvggt}     & Online  & 0.323             & \textbf{0.658}    & 0.059 & \textbf{0.972}    & 0.173             & 0.721 \\
OVGGT~\cite{lu2026ovggt}                 & Online  & 0.330             & 0.634             & \underline{0.058} & \underline{0.971} & 0.127             & 0.839 \\
CUT3R~\cite{wang2025cut3r}               & Online  & 0.459             & 0.476             & 0.072             & 0.950             & 0.141             & 0.829 \\
TTT3R~\cite{chen2025ttt3r}               & Online  & 0.432             & 0.497             & 0.063             & 0.969 & 0.136             & 0.851 \\
ZipMap~\cite{jin2026zipmap}              & Online  & \textbf{0.292}    & 0.606             & 0.077             & 0.955             & 0.096             & 0.920 \\
LingBot-Map~\cite{chen2026lingbotmap}    & Online  & 0.318             & \textbf{0.658}    & 0.059 & 0.960             & \textbf{0.070}    & \textbf{0.962} \\
\textbf{LinearVGGT (Ours)}               & Online  & \underline{0.317} & \underline{0.647} & \textbf{0.052}    & 0.969 & \underline{0.093} & \underline{0.947} \\
\bottomrule
\end{tabular}
\vspace{-1.0em}
\end{table}

%% file: table/cost.tex
\begin{table}[t]
\centering
\caption{\textbf{Training cost overview.} }
\label{tab:cost}
\footnotesize
\setlength{\tabcolsep}{4pt}
\renewcommand{\arraystretch}{0.92}
\begin{tabular}{l l r r}
\toprule
\textbf{Method} & \textbf{Training} & \textbf{GPU hours} & \textbf{Frames} \\
\midrule
VGGT~\cite{wang2025vggt}              & from scratch       & 13,824             & $\approx490$M   \\
LingBot-Map~\cite{chen2026lingbotmap} & from scratch       & $\approx$32,000    & $\approx520$M    \\
StreamVGGT~\cite{zhuo2025streamvggt}  & from VGGT          & 672                & 3.9M \\
\textbf{LinearVGGT (Ours)}            & from StreamVGGT    & \textbf{59}        & \textbf{1.2M} \\
\bottomrule
\end{tabular}
\vspace{-1.0em}
\end{table}

%% file: table/ablation_recon.tex
\begin{table}[t]
\centering
\vspace{-1.0em}
\caption{\textbf{Ablation study.} Each row removes or replaces one component of the full model. Mean and median accuracy error (Acc$\downarrow$) with dense views.}
\label{tab:ablation}
\footnotesize
\setlength{\tabcolsep}{3.5pt}
\renewcommand{\arraystretch}{0.92}
\begin{tabular}{@{}l l cc cc@{}}
\toprule
 & & \multicolumn{2}{c}{\textbf{7-Scenes}} & \multicolumn{2}{c}{\textbf{NRGBD}} \\
\cmidrule(lr){3-4} \cmidrule(l){5-6}
\textbf{Part} & \textbf{Setting} & Mean & Med. & Mean & Med. \\
\midrule
\multicolumn{2}{l}{\textbf{Full model}}                    & 0.037  & 0.020  & 0.048  & 0.024 \\
\addlinespace[2pt]
Layout & Mamba2 in all layers                    & 0.185 & 0.109  & 0.197 & 0.105 \\
\addlinespace[2pt]
Anchor & w/o anchor                                        & 0.049  & 0.024  & 0.055  & 0.027 \\
\addlinespace[2pt]
\multirow{2}{*}{Routing} & most recent frames              & 0.046  & 0.026  & 0.054  & 0.026 \\
 & random frames                                           & 0.079  & 0.042  & 0.050  & 0.024 \\
\addlinespace[2pt]
Indexer & w/o indexer (all tokens)        & 0.042  & 0.021  & 0.046  & 0.023 \\
\addlinespace[2pt]
Distill. & w/o ground-truth term                 & 0.106 &  0.071   & 0.078 & 0.056 \\
\bottomrule
\end{tabular}
\vspace{-1.0em}
\end{table}

%% file: tex/2_related.tex
\section{Related Works}

\label{sec:related work}

\paragraph{Transformer-based 3D Reconstruction Models}
Recent transformer-based 3D reconstruction models like VGGT, VGGT-$\Omega$ and $\pi^3$ have demonstrated that dense 3D geometry features can be predicted by transformers. However, these methods rely on dense global attention inference and this already become the bottleneck of 3D visual geometry transformers. While existing works use methods like sparse attention mechanisms or token -merging to reduce the cost of redundant global attention, their computational costs still increase quadratically with the number of images, hindering their applications to scenarios under large numbers of input images.
\vspace{-10pt}
\paragraph{Efficient Online 3D visual geometry transformers}
Robots~\cite{cadena2016slam}, AR/VR headsets~\cite{klein2007ptam,engel2023aria, maggio2025vggt-slam} and autonomous vehicles~\cite{bresson2017slamdriving,zuo2026dvgt} need the camera pose and the geometry of each frame as the frame arrives.
An offline model must process all frames again for each new frame, which is too slow for this setting. StreamVGGT~\cite{zhuo2025streamvggt} makes the global attention of VGGT causal and caches the keys and values of all past frames, so that a new frame reads the full history without recomputation. As a result, the cache and the cost of each frame grow with the stream, and the total cost grows quadratically. InfiniteVGGT~\cite{yuan2026infinitevggt} and XStreamVGGT~\cite{su2026xstreamvggt} prune this cache without retraining, and LongStream~\cite{cheng2026longstream} refreshes it periodically. LingBot-Map~\cite{chen2026lingbotmap} limits this context by position: a new frame attends to a few anchor frames, a window of recent frames and a trajectory memory of six tokens per past frame.
Recurrent models take the opposite approach. CUT3R~\cite{wang2025cut3r} reads from and writes to a fixed-size scene state at a constant cost per frame, and TTT3R~\cite{chen2025ttt3r} treats this state update as test-time training.
A fixed-size state, however, forgets earlier frames as the stream grows~\cite{chen2025ttt3r}.
ZipMap~\cite{jin2026zipmap} compresses all views into the fast weights of an MLP with large-chunk test-time training, which gives a linear-time bidirectional model with an extension to streaming.

\vspace{-10pt}
\paragraph{Linear Complexity Sequence Models}
Linear attention~\cite{katharopoulos2020transformers} writes causal attention as a recurrence over a fixed-size state, so each step has a constant cost and memory.
Mamba~\cite{gu2023mamba} and Mamba2~\cite{dao2024mamba2} add a selective, data-dependent decay to this state, and Gated DeltaNet~\cite{yang2025gdn} and Kimi Delta Attention~\cite{kimi2025linear} combine such a decay with the delta rule~\cite{schlag2021linear,yang2024deltanet}.
These models train in parallel with chunkwise kernels~\cite{yang2024fla} and run as recurrences at inference, so their total cost grows only linearly with the sequence length.
Their fixed-size state, however, recalls specific past tokens less exactly than attention~\cite{arora2024zoology,jelassi2024repeat} and often fails on sequences longer than those seen in training~\cite{benkish2025decimamba,ye2025longmamba,chen2024stuffed,buitrago2025length}, so hybrid models such as Jamba~\cite{lieber2024jamba} and Kimi Linear~\cite{kimi2025linear} keep a few attention layers.
In addition, a new architecture must be trained from scratch, which is costly for a large pretrained model such as VGGT.
Distillation avoids this cost by transferring the knowledge of a pretrained transformer into a linear-time model.
MOHAWK~\cite{bick2024mohawk} first matches the mixing matrices, then the hidden states of each block, and finally distills the full model, and Mamba in the Llama~\cite{wang2024mamballama} initializes Mamba layers from attention weights and keeps a quarter of the attention layers.
These methods target language models.
Inspired by MOHAWK, we propose a distillation method for visual geometry transformers.
It initializes the new layers from the attention weights of StreamVGGT, follows the three stages of MOHAWK, and adds ground-truth supervision to the last stage, so that the student can correct the errors of its teacher.

%% file: tex/7_conclusion.tex
\section{Conclusion}
We presented LinearVGGT, a streaming visual geometry transformer that runs at a nearly constant cost per frame.
Its Mamba2 layers summarize the stream in a state of fixed size, and its register-guided sparse attention layers restore what this summary forgets by reading a few relevant past frames.
Distilled from StreamVGGT with 9\% of the GPU hours that StreamVGGT itself required, LinearVGGT processes a 1,024-frame stream at 25 to 28\,ms per frame and outperforms its teacher in 3D reconstruction and video depth estimation.
These results suggest that a streaming visual geometry transformer does not need exact attention to its whole past: a compressed summary, together with direct access to a few frames chosen by their content, is enough.
We hope that our distillation makes it cheap to explore further recurrent designs for long streams.

%% file: tex/5_appendix.tex
\section{Implementation Details}
\label{sec:supp_impl}

This section gives the details that the main paper leaves out: the configuration and initialization of the recurrent layers (\cref{sec:supp_arch}), the training data (\cref{sec:supp_data}), the matrix mixer view behind the distillation (\cref{sec:supp_mixer}), the losses of the three stages (\cref{sec:supp_loss}), the optimization settings (\cref{sec:supp_hparams}) and the inference engine (\cref{sec:supp_fast}).

\subsection{Architecture}
\label{sec:supp_arch}

\paragraph{Mamba2 layers.}
Each of the 18 recurrent layers is an unmodified Mamba2 block~\cite{dao2024mamba2} whose shape follows the teacher attention it replaces: width 1024, 16 heads of dimension $d = 64$, a state of size $n = 64$ per head, no expansion, and a causal convolution with a kernel of 4 tokens.
The block projects each token $u_i$ with one linear map into a gate $\mathbf{z}_i$, the input $\mathbf{x}_i$, the vectors $\mathbf{b}_i$ and $\mathbf{c}_i$ of every head, and one step logit $\delta_i$ per head.
The convolution and a SiLU act on $\mathbf{x}_i$, $\mathbf{b}_i$ and $\mathbf{c}_i$, and the step is $\Delta_i = \operatorname{softplus}(\delta_i + \beta)$ with a learned bias $\beta$.
With the two terms that \cref{eq:ssm} omits, the output of the layer is
\begin{equation}
  \mathbf{o}_i = W_o \operatorname{RMSNorm}\!\big( (H_i \mathbf{c}_i + D\, \mathbf{x}_i) \odot \operatorname{SiLU}(\mathbf{z}_i) \big),
  \label{eq:supp_mamba_out}
\end{equation}
where $D$ is a learned skip weight per head and $W_o$ the output projection.
The state of one layer holds $16 \times 64 \times 64$ numbers and its convolution buffer the last three tokens, so the 18 layers together carry 1.2M numbers from frame to frame, which is 4.5\,MiB in single precision.

\paragraph{Initialization from the teacher.}
Each Mamba2 layer starts from the weights $W_Q$, $W_K$, $W_V$ and $W_O$ of its teacher layer:
\begin{equation}
\begin{aligned}
  \mathbf{x}_i &\leftarrow W_V u_i, & \mathbf{b}_i &\leftarrow W_K u_i, \\
  \mathbf{c}_i &\leftarrow \tfrac{1}{\sqrt{d}}\, W_Q u_i, & W_o &\leftarrow W_O ,
\end{aligned}
\label{eq:supp_init}
\end{equation}
with the biases copied in the same way.
The remaining parameters are set so that the new parts of the block start as identities.
The convolution passes its last tap unchanged, the skip weight $D$ is zero because attention has no skip path, and the gate projection is zero with a constant bias, which the RMSNorm after it cancels.
The step projection is also zero, so every token starts with the same step $\Delta = \ln 2 / (2N)$ at $a = -1$, and the state halves over the $2N = 2082$ tokens of two frames.
Because $\Delta$ also scales the input in \cref{eq:ssm}, the initial state is a normalized running average of the past and not an unbounded sum.
The query--key normalization and the rotary embedding of the teacher have no counterpart in the block and are dropped.

\paragraph{Token-causal scan.}
The teacher is causal over frames: a token attends to all tokens of its own frame, including those after it in raster order.
The scan of a recurrent layer is causal over tokens, so a token cannot read the later tokens of its frame in this layer.
The frame attention layers of the same block, which we keep, see all tokens of a frame at once and supply this context.

\paragraph{Sparse attention layers.}
The six sparse layers reuse the teacher attention module, and the routing of \cref{eq:route} takes its register queries and keys after the query--key normalization and the rotary embedding of this module.
The softmax of \cref{eq:route} runs over the register keys of all frames up to the current one, and the top-$k$ step then ranks the frames $1, \dots, t{-}1$, with ties resolved in favor of the earlier frame.
While $t \le k + 1$, the layer reads every past frame.
For token eviction, the score of patch $j$ of frame $s$ is the attention that it receives inside its own frame,
\begin{equation}
  e_{s,j} = \sum_{h=1}^{H} \sum_{i=1}^{N} \operatorname{softmax}_{j'}\!\Big(\tfrac{1}{\sqrt{d}}\, \mathbf{q}^{h\top}_{s,i}\, \mathbf{k}^h_{s,j'}\Big)_{j} ,
  \label{eq:supp_evict}
\end{equation}
where the softmax runs over the $N$ tokens of frame $s$.
With $\rho = 1/4$, a past frame keeps its 5 special tokens and 259 of its 1036 patches, so frame $t$ attends to $2N + 3 \cdot 264 = 2874$ keys in a sparse layer.
Selection and eviction are active during training with the same $k$ and $\rho$ as at test time.
Training uses a full-sequence form of the layer in which every frame of a clip gathers its selected frames in parallel, and it returns the same output as the streaming form with a cache.

\subsection{Training data}
\label{sec:supp_data}

\begin{table}[H]
\centering
\caption{\textbf{Training data.} Weight: share of the training clips. Scenes: scenes, sequences or objects that we use. Max.\ gap: largest interval between two consecutive frames of a clip, in stored frames. Ordered: share of the clips that keep the temporal order; the frames of the other clips are shuffled.}
\label{tab:supp_data}
\small
\setlength{\tabcolsep}{3pt}
\begin{tabular}{llrrrr}
\toprule
 & Dataset & \shortstack[r]{Weight\\(\%)} & Scenes & \shortstack[r]{Max.\\gap} & \shortstack[r]{Ordered\\(\%)} \\
\midrule
\multirow{4}{*}{\rotatebox{90}{Indoor}}
  & ScanNet ~\cite{dai2017scannet}             & 18.1 &  705 &   3 &  60 \\
  & ARKitScenes ~\cite{baruch2021arkitscenes}        & 18.1 & 1477 &   3 &  80 \\
  & ARKitScenes-HighRes &  9.0 &  250 &   3 &  50 \\
  & HyperSim  ~\cite{roberts2021hypersim}          &  4.8 &  149 &   4 &  50 \\
\midrule
\multirow{3}{*}{\rotatebox{90}{Object}}
  & CO3Dv2  ~\cite{reizenstein2021co3d}            &  9.3 &  608 &  12 &  50 \\
  & WildRGB-D  ~\cite{xia2024wildrgbd}         &  9.3 &  368 &   6 &  50 \\
  & OmniObject3D ~\cite{wu2023omniobject3d}       &  3.7 &  501 & 100 &   0 \\
\midrule
\multirow{4}{*}{\rotatebox{90}{Outdoor}}
  & BlendedMVS ~\cite{yao2020blendedmvs}         &  4.6 &  113 &   8 &   0 \\
  & MegaDepth  ~\cite{li2018megadepth}         &  4.6 &  195 &   8 &   0 \\
  & Waymo ~\cite{sun2020waymo}              &  4.6 &  200 &   4 &  90 \\
  & Virtual KITTI 2 ~\cite{cabon2020vkitti2}    &  4.6 &   50 &   5 & 100 \\
\midrule
\multirow{3}{*}{\rotatebox{90}{Synth.}}
  & Spring  ~\cite{mehl2023spring}            &  3.7 &   36 &  16 & 100 \\
  & MVS-Synth ~\cite{huang2018deepmvs}           &  0.9 &  120 &   4 & 100 \\
  & PointOdyssey~\cite{zheng2023pointodyssey}        &  4.6 &   33 &   1 & 100 \\
\bottomrule
\end{tabular}
\end{table}

\paragraph{Datasets and mixture.}
We use 14 of the 15 training datasets of StreamVGGT, all except ScanNet++, and a subset of the scenes of each (\cref{tab:supp_data}).
Indoor scenes receive half of the training clips.
Inside the indoor group, and across the object, outdoor and synthetic groups that share the other half, the weights keep the proportions of StreamVGGT.
None of the evaluation datasets (7-Scenes, NRGBD, TUM, Sintel, Bonn, KITTI) is used for training.

\paragraph{Clip sampling.}
Clips are sampled with the rule of CUT3R~\cite{wang2025cut3r}.
A clip starts at a random frame of a scene, and the interval between two consecutive frames is drawn uniformly between 1 and a maximum that depends on the dataset.
With a probability that also depends on the dataset, the clip keeps its temporal order, and a part of these ordered clips uses one interval for all frames.
The frames of the other clips are shuffled, so the student also sees unordered image sets with large changes of view.
We take the probabilities from CUT3R and convert its maximum intervals to the frame rates at which we store each dataset (\cref{tab:supp_data}).
From 160 candidate clips per scene, we draw 76,000 clips according to the weights of \cref{tab:supp_data}, and Stage~3 uses each of them once.

\paragraph{Preprocessing.}
Every frame is resized to cover $518{\times}392$ pixels and cropped at its center; clips from portrait sources use $392{\times}518$.
We apply no color or geometric augmentation.

\subsection{Matrix mixers}
\label{sec:supp_mixer}

The distillation of the main paper rests on the matrix mixer view of sequence layers~\cite{dao2024mamba2,bick2024mohawk}.
We summarize this view here and state what it does and does not guarantee for our student.

\paragraph{Sequence layers as matrices.}
Consider one head of a layer that maps the $T$ tokens $u_1, \dots, u_T$ of a clip, stacked in $U$, to $T$ outputs.
Attention and the recurrent layers that we use share the form
\begin{equation}
  Y = M(U)\, X, \qquad X = U W_x ,
  \label{eq:supp_mixer}
\end{equation}
where $X \in \mathbb{R}^{T \times d}$ holds one value per token and $M(U) \in \mathbb{R}^{T \times T}$ is a mixing matrix that the layer computes from its input.
Row $i$ of $M$ states how much of the value of every token the output $i$ receives.
Layers of this form differ only in how they parameterize $M$ and in what the product $M X$ costs.
For the global attention of the teacher (\cref{eq:global}), the matrix is dense,
\begin{equation}
  M = A = \operatorname{softmax}\!\big(\tfrac{1}{\sqrt{d}}\, Q K^\top + B_{\text{causal}}\big),
  \label{eq:supp_attn_mixer}
\end{equation}
with a mask $B_{\text{causal}}$ that hides the tokens of later frames, and the product costs $\mathcal{O}(T^2)$.
For a Mamba2 layer, unrolling the recurrence gives the matrix of \cref{eq:mixing}, $M_{ij} = \mathbf{c}_i^\top \mathbf{b}_j \exp(a \sum_{m=j+1}^{i} \Delta_m)\, \Delta_j$ for $j \le i$ and zero otherwise.
This matrix is the elementwise product of a term of rank $n$, $\mathbf{c}_i^\top \mathbf{b}_j$, and a decay mask, and this structure is what lets the recurrence of \cref{eq:ssm} compute $M X$ in $\mathcal{O}(T)$ without ever forming $M$.
Attention and Mamba2 are thus two parameterizations of the same object, one free and quadratic, one structured and linear.

\paragraph{Why matching the matrix transfers the layer.}
A global block of the teacher consists of a layer normalization, the mixer with its output projection, LayerScale, a residual connection and an MLP.
The student copies all of these parts except the mixing matrix, and \cref{eq:supp_init} also gives it the value and output projections of the teacher.
For the same input $U$, a teacher head then returns $A V$ and the student head $M X$ with $X = V$, so that
\begin{equation}
  \| M X - A V \|_F \;\le\; \| M - A \|_F\, \| V \|_F .
  \label{eq:supp_bound}
\end{equation}
The error of the layer output is bounded by the error of the mixing matrix, and a student layer with $M = A$ returns exactly the output of its teacher layer.
All parts after the mixer are copies, so the same holds for the output of the block and, layer after layer, for the predictions of the network.
The distillation of a 1.28B-parameter model thus reduces to the regression of one $T \times T$ matrix per head and layer.
These problems are independent across layers once every layer receives the input of its teacher layer, which is the setting of Stage~1.
The sparse attention layers are the trivial case: they compute the teacher matrix itself on the selected frames and need no matrix stage.

\paragraph{Why the matrix stage is not enough.}
Two facts keep Stage~1 from finishing the distillation.
First, $M = A$ cannot be reached.
The Mamba2 matrix is a rank-$n$ matrix with $n = 64$ under a decay mask and is causal over tokens, while the attention matrix is unconstrained and causal over frames, so a residual $\| M - A \|_F > 0$ remains in every layer.
Second, we keep the official Mamba2 block, in which a SiLU follows the convolution on $X$ and an RMSNorm precedes the output projection (\cref{eq:supp_mamba_out}); MOHAWK removes both from its student.
With them, $X = V$ and \cref{eq:supp_bound} hold only approximately, even at initialization.
Stage~2 therefore trains each layer on its output, which includes these two parts, and Stage~3 trains all layers together, because the teacher inputs of the first two stages hide how the residual errors of 24 layers accumulate.

\subsection{Losses}
\label{sec:supp_loss}

\paragraph{Stage 1.}
The set $\Omega$ contains 96 query tokens at even spacing over the $8N$ tokens of a clip, and the loss compares the corresponding rows of the mixing matrix and of the teacher attention in all 16 heads.
The teacher rows are the softmax of \cref{eq:global} under the causal mask over frames.
A teacher row therefore has mass on the later tokens of its own frame, which the token-causal mixing matrix cannot reproduce; this part of the target sets a floor for $\mathcal{L}_1$ and is recovered by the frame attention layers.
The output projection and the RMSNorm of \cref{eq:supp_mamba_out} do not enter the mixing matrix and stay fixed in this stage.

\paragraph{Stage 2.}
Every Mamba2 layer receives the input of its teacher layer and is trained on the relative error of its output.
The 18 layers are independent in this stage, so we keep for each layer the weights with the lowest error on the held-out clips and start Stage~3 from them.

\paragraph{Distillation term of Stage 3.}
Let $\hat{D}, \hat{P}, \hat{\mathbf{t}}, \hat{\mathbf{q}}, \hat{\mathbf{f}}$ be the depth maps, point maps, camera translations, rotation quaternions and fields of view that the student predicts for a clip, and $D, P, \mathbf{t}, \mathbf{q}, \mathbf{f}$ those of the teacher.
With the relative error $r(\hat{x}, x) = \operatorname{mean}(\|\hat{x} - x\|^2) / \operatorname{mean}(\|x\|^2)$ over all frames and pixels of the clip,
\begin{equation}
\begin{aligned}
  \mathcal{L}_{\text{KD}} = \tfrac{1}{5} \Big[\, & r(\hat{D}, D) + r(\hat{P}, P) + r(\hat{\mathbf{f}}, \mathbf{f}) \\
  & + \frac{\operatorname{mean}\|\hat{\mathbf{t}} - \mathbf{t}\|^2}{\max\!\big(\operatorname{mean}\|\mathbf{t}\|^2,\ (0.05\, \sigma_P)^2\big)} \\
  & + \operatorname{mean}\, \min\!\big(\|\hat{\mathbf{q}} - \mathbf{q}\|^2, \|\hat{\mathbf{q}} + \mathbf{q}\|^2\big) \Big],
\end{aligned}
\label{eq:supp_kd}
\end{equation}
where $\sigma_P$ is the root mean square of the teacher point map.
The floor in the translation term matters for clips in which the camera hardly moves: there the mean squared translation of the teacher is close to zero and the plain relative error has no bound.

\paragraph{Ground-truth term of Stage 3.}
The ground-truth point map $P^\star$ is the ground-truth depth $D^\star$ unprojected with the ground-truth intrinsics and moved into the coordinate frame of the first camera of the clip.
Prediction and ground truth are each divided by their own scale before they are compared: depths by the mean depth, $\bar{d}$ and $\bar{d}^\star$, and point maps and translations by the mean distance of the points to the origin, $\bar{p}$ and $\bar{p}^\star$, all taken over the set $\mathcal{V}$ of pixels with valid ground truth in the clip.
The term is the mean of five parts,
\begin{equation}
\begin{aligned}
  \mathcal{L}_{\text{GT}} = \tfrac{1}{5} \Big[\, & \operatorname*{mean}_{\mathcal{V}} \Big| \tfrac{\hat{D}}{\bar{d}} - \tfrac{D^\star}{\bar{d}^\star} \Big| + \operatorname*{mean}_{\mathcal{V}} \min\!\Big( \Big\| \tfrac{\hat{P}}{\bar{p}} - \tfrac{P^\star}{\bar{p}^\star} \Big\|,\, 1 \Big) \\
  & + \operatorname{mean} \Big| \tfrac{\hat{\mathbf{t}}}{\bar{p}} - \tfrac{\mathbf{t}^\star}{\bar{p}^\star} \Big| + \operatorname{mean} \big| \hat{\mathbf{f}} - \mathbf{f}^\star \big| \\
  & + \operatorname{mean}\, \min\!\big(\|\hat{\mathbf{q}} - \mathbf{q}^\star\|^2, \|\hat{\mathbf{q}} + \mathbf{q}^\star\|^2\big) \Big],
\end{aligned}
\label{eq:supp_gt}
\end{equation}
with fields of view in radians.
Because $\mathcal{L}_{\text{GT}}$ does not depend on the scale of the prediction, it does not compete with $\mathcal{L}_{\text{KD}}$, which alone fixes the output scale of the student to that of the teacher.
A frame contributes to $\mathcal{L}_{\text{GT}}$ only if its ground-truth depth covers at least half of the image; the threshold is 2\% for OmniObject3D, whose objects lie on an empty background, and 3\% for MegaDepth and Waymo, whose depth is sparse.
The three pose parts need a valid pose for the frame and for the first frame of the clip, and a part without any valid frame is left out of the mean.
We also drop the HyperSim frames whose stored pose does not reproduce the rendered positions to a relative error of $10^{-2}$.

\subsection{Optimization}
\label{sec:supp_hparams}

\begin{table}[H]
\centering
\caption{\textbf{Training settings of the three stages.}}
\label{tab:supp_hparams}
\small
\setlength{\tabcolsep}{3.5pt}
\begin{tabular}{lccc}
\toprule
 & Stage 1 & Stage 2 & Stage 3 \\
\midrule
Target & attention & layer output & predictions \\
Layer inputs from & teacher & teacher & student \\
Trained layers & 18 Mamba2 & 18 Mamba2 & all 24 \\
Steps & 2,000 & 4,000 & 38,000 \\
Learning rate & $10^{-4}$ & $10^{-4}$ & $10^{-5}$ \\
Warm-up steps & -- & -- & 200 \\
Cosine decay & -- & -- & last 3,800 steps \\
Frames per clip & 8 & 8 & 8 / 16 / 24 \\
\midrule
Optimizer & \multicolumn{3}{c}{AdamW, $\beta = (0.9, 0.999)$, no weight decay} \\
Clips per step & \multicolumn{3}{c}{2} \\
Gradient clipping & \multicolumn{3}{c}{norm 1.0} \\
Precision & \multicolumn{3}{c}{bf16 aggregator, fp32 heads} \\
Input size & \multicolumn{3}{c}{$518{\times}392$} \\
\bottomrule
\end{tabular}
\end{table}

\Cref{tab:supp_hparams} lists the settings of the three stages.
Stage~3 has two phases.
The first phase updates only the new layers: the six sparse layers train together with the 18 Mamba2 layers, while the encoder, the frame attention layers, the MLPs and normalization layers of the global blocks and the heads stay frozen, as in Stages~1 and~2.
In this phase, the clips have 8 frames for the first 22,800 steps, 16 frames for the next 7,600 and 24 frames for the last 7,600, which gives about 0.97M frames.
The second phase then updates all layers except the image encoder for a further 0.13M frames, so that the frame attention layers and the heads adapt to the new cross-frame layers.
Together with the 0.1M frames of Stages~1 and~2, the whole distillation uses 1.2M frames.
The two clips of a step are processed one after the other and their gradients are accumulated, with gradient checkpointing in the aggregator blocks, so that a 24-frame clip fits on one GPU.
We evaluate the student every 1,000 steps on the 140 held-out clips and keep the checkpoint with the lowest $\mathcal{L}_{\text{GT}}$.

\subsection{Efficient streaming inference}
\label{sec:supp_fast}

\begin{table}[H]
\centering
\caption{\textbf{Where FastStreamer saves time.} One streaming frame of LinearVGGT, averaged over a 48-frame stream. Wall: latency. Host: time on the CPU side. Kernels: GPU kernels launched per frame.}
\label{tab:supp_fast}
\small
\setlength{\tabcolsep}{4pt}
\begin{tabular}{l rrr c}
\toprule
 & \shortstack[r]{Wall\\(ms)} & \shortstack[r]{Host\\(ms)} & Kernels & \shortstack[c]{Bitwise\\identical} \\
\midrule
Base & 34.8 & 17.9 & 4,235 & reference \\
+ exact optimizations & 29.7 & 16.4 & 3,155 & yes \\
+ fused kernels & 26.3 & 16.2 & 1,939 & no \\
\bottomrule
\end{tabular}
\end{table}

\paragraph{CUDA graphs.}
In eager execution, a frame launches more than 4,000 GPU kernels, and the Python and dispatch work between them takes most of the 77.2\,ms.
FastStreamer writes the computation of one frame as a single program and cuts it at the six sparse attention cores, whose key and value caches grow with the stream.
Each of the seven segments between the cuts has fixed tensor shapes and is captured once per resolution as a CUDA graph, with the states and convolution buffers of the Mamba2 layers in static buffers that the graph updates in place.
The first segment holds the encoder and the layers before the first sparse layer, and the last one ends with the DPT head.
The sparse attention cores and the camera head, which has a growing cache of its own, run eagerly between the graphs.
This step alone brings the latency from 77.2 to 34.8\,ms without any change of the outputs.

\paragraph{Exact optimizations.}
Five further changes remove repeated work and leave every output bit unchanged.
(i)~The linear and convolution weights of the aggregator are stored in bf16, the precision to which autocast converts them on every call.
(ii)~The positional embeddings of the encoder and of the DPT head are computed once per resolution and not once per frame.
(iii)~The rotary tables are gathered once per resolution and not in every attention call.
(iv)~Only the intermediate features that the heads read are assembled.
(v)~The sparse layers cache their keys after query--key normalization and rotary embedding, so a frame normalizes and rotates only its own keys and not those of all selected frames.

\paragraph{Fused kernels.}
Two Triton kernels then replace chains of small kernels: one computes the residual addition, LayerScale and the following layer normalization, and one the query--key normalization with the rotary embedding.
They halve the number of kernels of the remaining program and bring the latency to 26.3\,ms.
Their layer normalization statistics differ from those of PyTorch at the level of fp32 rounding, so the outputs are close to the eager ones but not bitwise identical.
On a 200-frame TUM sequence, the median relative difference per frame is $1.5 \times 10^{-3}$ for depth, $4.0 \times 10^{-3}$ for point maps and $4.5 \times 10^{-4}$ for the camera encoding.
For comparison, a switch between two attention backends of PyTorch in the eager model changes the same outputs by $1.4 \times 10^{-3}$, $4.1 \times 10^{-3}$ and $4.7 \times 10^{-4}$, so the fused kernels stay within the numerical noise of the eager path.

\FloatBarrier
\section{Evaluation Details}
\label{sec:supp_eval}

\paragraph{3D reconstruction.}
We use the 18 test sequences of 7-Scenes and the 9 scenes of NRGBD with the split and the code of CUT3R and StreamVGGT.
With \emph{sparse views}, the input is one keyframe every 200 frames on 7-Scenes and every 500 frames on NRGBD; with \emph{dense views}, one every 20 and 50 frames, which gives 8 to 50 views per sequence.
For the long streams of \cref{sec:recon}, we take every second frame of a 7-Scenes sequence and keep the first $N$ of them.
The predicted point maps of all views are cropped to their central $224{\times}224$ pixels, aligned to the ground truth with a similarity transform from the pixel correspondences~\cite{umeyama1991least} and refined with ICP at a distance threshold of 0.1\,m.
Accuracy is the distance from a predicted point to its nearest ground-truth point, completeness the distance from a ground-truth point to its nearest predicted point, and normal consistency the agreement between the normals of matched points.
For every sequence, we take the mean and the median of each quantity over its points, and the tables report the average of both over the sequences (Mean and Med.).

\paragraph{Video depth.}
We follow CUT3R on Sintel, Bonn and KITTI and feed each sequence as one stream.
The predicted depth of a sequence is aligned to the ground truth with one scale factor, as in \cref{tab:vdepth}, or with a scale and a shift, as in \cref{tab:supp_vdepth}.

\paragraph{Camera pose.}
We follow CUT3R on Sintel (14 sequences), on the 8 dynamic sequences of TUM~\cite{sturm2012tumrgbd} and on 100 test scenes of ScanNet, the latter two with the first 90 frames at a stride of 3.
The predicted trajectory is aligned to the ground truth with a similarity transform, and we report the absolute trajectory error (ATE) and the relative pose errors in translation and rotation.

\paragraph{Efficiency measurement.}
We time all methods on one RTX PRO 6000 Blackwell GPU with batch size 1 and $518{\times}392$ frames.
The latency of a frame covers the full step from the image to the outputs of all heads and is measured with CUDA events; memory is the peak allocated GPU memory.

\FloatBarrier
\section{Additional Results}
\label{sec:supp_results}

This section reports the results of the final model that the main paper leaves out: reconstruction with dense views (\cref{sec:supp_dense}), video depth with a scale and a shift (\cref{sec:supp_vdepth}), camera pose (\cref{sec:supp_pose}), the results for every test sequence (\cref{sec:supp_per_scene}) and an analysis of the frame routing (\cref{sec:supp_routing}).

\subsection{Reconstruction with dense views}
\label{sec:supp_dense}

\input{table/tab_recon_dense}

\Cref{tab:supp_recon_dense} repeats the comparison of \cref{tab:recon} with dense views.
LinearVGGT again improves on its teacher on both datasets and nearly halves its accuracy error on NRGBD (0.048 against 0.088).
On NRGBD, it is also more accurate than CUT3R and TTT3R, while on 7-Scenes these two models have a lower error; LingBot-Map and streaming ZipMap are more accurate on both datasets.
With many overlapping views, the offline models, which see all frames at once, are more accurate than every online model.

\subsection{Video depth with scale and shift}
\label{sec:supp_vdepth}

\input{table/tab_vdepth_scale_shift}

With a scale and a shift per sequence (\cref{tab:supp_vdepth}), the ranking of \cref{tab:vdepth} does not change.
LinearVGGT has the lowest absolute relative error of the online models on Bonn and the second lowest on Sintel and KITTI, and on KITTI it lowers the error of its teacher from 0.164 to 0.074.

\subsection{Camera pose}
\label{sec:supp_pose}

\input{table/tab_pose}

As shown in \cref{tab:supp_pose}, LinearVGGT trails top online baselines such as LingBot-Map and StreamVGGT in global trajectory accuracy (ATE), yet remains competitive on local relative pose metrics—ranking third among online methods in $\text{RPE}_r$ on both TUM-dynamics ($0.371^\circ$) and ScanNet ($0.587^\circ$). This gap may come from the mixture of our training data.
We consider this the most likely cause and the main limitation of the current model.

\subsection{Results per sequence}
\label{sec:supp_per_scene}

\input{table/tab_per_scene}

\Cref{tab:supp_per_scene} lists the accuracy and completeness of StreamVGGT and LinearVGGT for every test sequence.
With sparse views, LinearVGGT is more accurate than its teacher on 16 of the 18 sequences of 7-Scenes and on 5 of the 9 scenes of NRGBD; with dense views, on 14 of 18 and on 8 of 9.
The table also shows why we report medians next to means.
With sparse views, a model occasionally misregisters one of the few input frames, and the error of this sequence then exceeds that of a typical sequence by an order of magnitude.
StreamVGGT does so on three sequences of 7-Scenes (redkitchen/seq-04, office/seq-06 and office/seq-09, with errors of 0.39 to 0.49), and LinearVGGT recovers two of them, which accounts for about 60\% of the difference between the two means.
On NRGBD, the scene kitchen has an error of about 0.3 for both models and dominates both means.
The gain of LinearVGGT on the other sequences is smaller but consistent, as the count of sequences and the medians of \cref{tab:recon} show.

\subsection{Frame routing}
\label{sec:supp_routing}

\input{table/tab_routing}

The routing of \cref{eq:route} ranks the past frames with the register tokens alone.
We compare it with an oracle that ranks them by the attention mass that all tokens of the new frame give to all cached tokens of a past frame, and that selects the $k=3$ frames with the most mass.
We report the agreement, the share of the selected frames that the oracle also selects, and the recovered mass, the mass on the selected frames relative to the mass on the oracle frames.
We stream the eight TUM-dynamics sequences (90 frames) and the first 500 frames of two of them.

In \cref{tab:supp_routing}a, the registers select an oracle frame for 52\% of their picks on the 90-frame streams and for 44\% on the 500-frame streams, against 12\% and 3\% for random frames, and they recover 96\% of the oracle mass on both.
The agreement is far from complete while almost no mass is lost, so the frames on which the two selections differ carry nearly the same mass.
\Cref{tab:supp_routing}b resolves this by layer.
Layers 11 and 15 have a clear preference: their oracle frames hold 2.5 and 2.0 times the mass of random frames, and there the registers agree with the oracle for 100\% and 74\% of their picks.
In the other four layers, the oracle frames hold at most 1.2 times the mass of random frames, so the ranking is nearly arbitrary and an agreement of 14\% to 36\% costs less than 9\% of the mass.

\FloatBarrier
\section{More Qualitative Results}
\label{sec:supp_qualitative}

\begin{figure}[t]
\centering
\includegraphics[width=\linewidth]{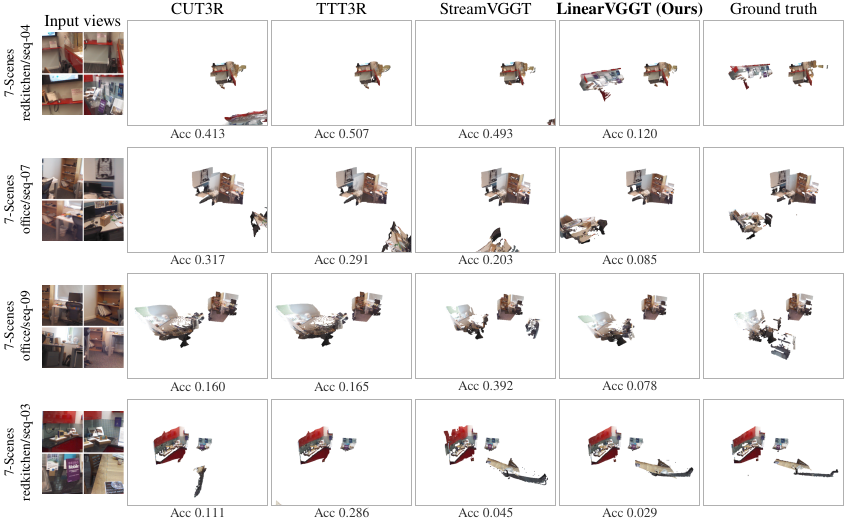}
\caption{\textbf{Reconstruction from sparse views with little overlap.} Four sequences of 7-Scenes with five input views each (four are shown). Every point cloud is the output of a method after the alignment of the benchmark, rendered from the same virtual camera, with the accuracy error of the sequence below it. In the first three rows, the baselines place at least one view far from its true position, in several cases outside the frame, while LinearVGGT places it close to the ground truth.}
\label{fig:supp_qualitative}
\end{figure}

\Cref{fig:supp_qualitative} shows reconstructions of four sequences of 7-Scenes from sparse views.
We selected them as examples of the case that separates the methods in \cref{tab:supp_per_scene}: one of the input views shares almost no content with the others, so its place in the scene must come from the geometry that the model has built up and not from matched image content.
In redkitchen/seq-04 and office/seq-07, CUT3R, TTT3R and StreamVGGT all reconstruct the view that looks at a different part of the room with a plausible shape but at a wrong position, which raises their accuracy error to between 0.2 and 0.5.
LinearVGGT places this view next to the others as in the ground truth.
In office/seq-09, the two recurrent models distort the largest surface and StreamVGGT detaches a part of the scene, and in redkitchen/seq-03 the recurrent models misplace the view of the table.
These examples are favorable to LinearVGGT by selection, and \cref{tab:supp_per_scene} gives the errors of all sequences; office/seq-06, for instance, is a sequence on which every method, including ours, misplaces a view.

\FloatBarrier
\section{More on the Attention of the Teacher}
\label{sec:supp_attention}

\Cref{fig:redundancy} shows, for every global layer of StreamVGGT, how many past frames and tokens hold 90\% of the attention of a new frame.
This section looks at the same attention in more detail, and each of its observations corresponds to one element of our sparse attention: the anchor, the routing that is shared by all tokens, and the token indexer.

\begin{figure}[t]
\centering
\includegraphics[width=0.6\linewidth]{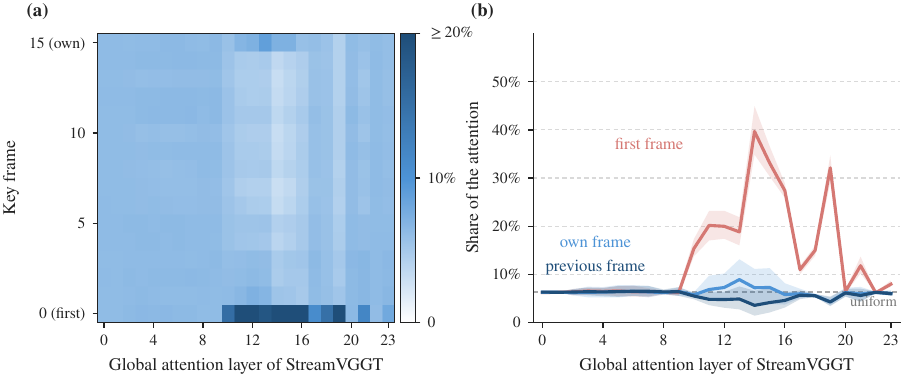}
\caption{\textbf{The first frame is the one past frame that the teacher singles out.} (a) Share of the attention of a new frame that goes to each of the 16 frames of a clip, for every global layer of StreamVGGT. (b) The same share for the first frame, the previous frame and the new frame itself (mean and 10th to 90th percentile over the clips). With uniform attention, every frame receives 6.25\%.}
\label{fig:supp_attn_frames}
\end{figure}

\begin{figure}[t]
\centering
\includegraphics[width=0.8\linewidth]{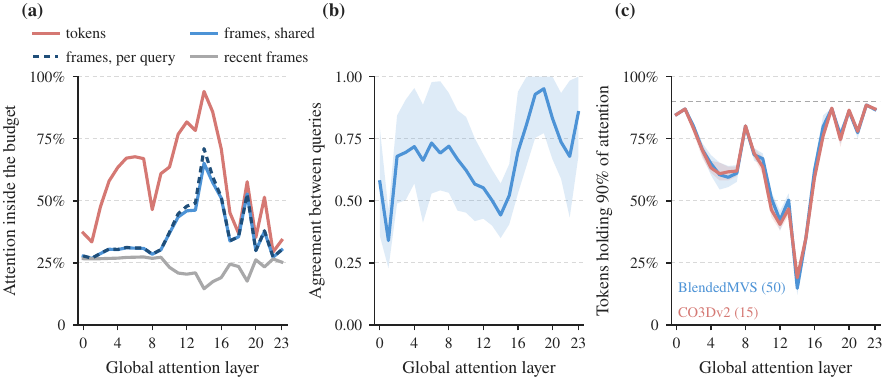}
\caption{\textbf{What a small budget of past frames or tokens can hold.} (a) Share of the attention to the past that lies inside a budget of four of the 15 past frames: the four best frames chosen once for all queries (shared) or by every query, the same number of individual tokens, and the four most recent frames. Uniform attention gives 27\%. (b) Overlap (Jaccard index) between the three best frames of two queries of the same frame. (c) Share of the past tokens that holds 90\% of the attention, for the clips of each dataset. Bands show the 10th to 90th percentile over the clips.}
\label{fig:supp_attn_budget}
\end{figure}

\begin{figure}[t]
\centering
\includegraphics[width=\linewidth]{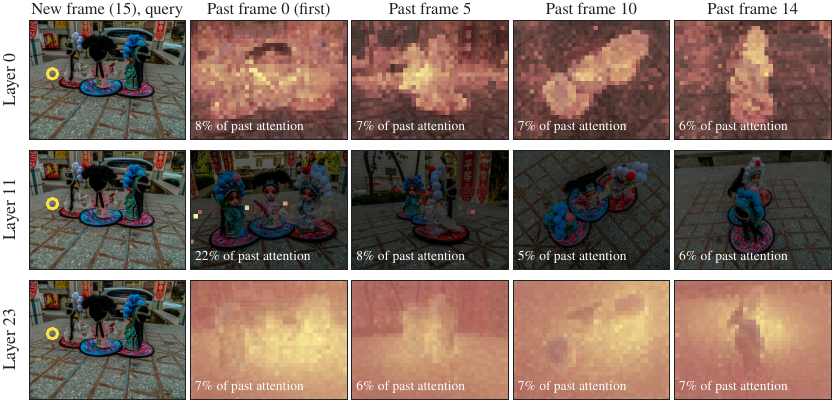}
\caption{\textbf{Attention of one patch of a new frame to four past frames.} The query patch is marked in the left column. Each row shows one global layer of StreamVGGT, with its own color scale, and each past frame is labeled with its share of the attention that the query gives to the past. Layers 0 and 23 spread their attention over all frames and most of their patches; layer 11 places it on a few patches, most of them in the first frame.}
\label{fig:supp_attn_spatial}
\end{figure}

\paragraph{Setup.}
We run StreamVGGT in streaming mode on clips of 16 frames from BlendedMVS and CO3Dv2 and record the attention of the last frame in all 16 heads of every global layer.
After the removal of clips with corrupted frames, 65 clips remain, 50 from BlendedMVS and 15 from CO3Dv2.
The queries are an $8{\times}8$ grid of patch tokens and the five special tokens of the last frame, and the keys are all tokens of the 16 frames.
Between 87\% and 93\% of the attention of a query goes to the patch tokens of the 15 past frames, and at most 8\% to special tokens.

\paragraph{The first frame stands out.}
\Cref{fig:supp_attn_frames} shows how the attention of a new frame is divided among the frames of the clip.
In layers 0 to 9, every frame receives the uniform share of about 6\%.
From layer 10 on, the first frame receives far more: 15\% to 20\% in layers 10 to 13, 40\% in layer 14, and still 32\% in layer 19.
No other frame is treated in this way.
The previous frame, which is closest in time and usually in viewpoint, receives at most its uniform share and less than that in layers 10 to 19, and the new frame itself receives 4\% to 9\%.
The teacher thus returns to the frame that defines the coordinate system in most of its upper layers, which is why the sparse layers of LinearVGGT always read the first frame as an anchor.

\paragraph{One set of frames serves all tokens.}
Our routing selects the same past frames for all tokens of a new frame.
To see what this costs, we compare the attention inside the four best past frames when every query chooses its own frames and when one choice is shared by all queries (\cref{fig:supp_attn_budget}a).
The two curves nearly coincide: they differ by at most 0.06, in layer 14.
The reason is that the queries agree on the frames: the three best frames of two queries of the same frame overlap with a Jaccard index of 0.34 to 0.95, and of at least 0.5 in 22 of the 24 layers (\cref{fig:supp_attn_budget}b).
A selection per frame therefore loses little against a selection per token.
The budget itself is small, however: outside layers 10 to 19, the four best frames hold 27\% to 31\% of the attention, hardly more than four random frames, so in these layers no choice of frames is better than another.
The same panel shows that recency is a poor way to choose: in layers 10 to 23, the four most recent frames hold less attention than four random frames would, and in layer 14 they hold 15\% where the four best frames hold 65\%.

\paragraph{Inside a frame, a few tokens carry the mass.}
Four of 15 frames contain 27\% of the past tokens.
If a query may spend this budget on individual tokens anywhere in the past, it captures far more attention than with four whole frames: 94\% against 65\% in layer 14, and 77\% against 44\% in layer 11 (\cref{fig:supp_attn_budget}a).
The attention inside the selected frames is therefore concentrated on a subset of their tokens.
This motivates the token indexer but does not validate it: the curve lets every query choose its own tokens, and in the concentrated layers 11 to 15 the 64 best tokens of two queries overlap with a Jaccard index of only 0.02 to 0.06.
The indexer keeps one subset per frame for all later queries, so it cannot recover this per-query mass, and its justification is the ablation of \cref{tab:supp_sparse_hparams}, not this statistic.
\Cref{fig:supp_attn_spatial} shows an example: in layer 11, a query patch on the ground attends to a handful of patches of the first frame and hardly to anything else, while in layers 0 and 23 its attention covers all frames and most of their patches.

\paragraph{The pattern holds across datasets.}
The two datasets give nearly the same curve (\cref{fig:supp_attn_budget}c).
In layer 14, 90\% of the attention lies on 15\% of the past tokens for the scenes of BlendedMVS and on 19\% for the objects of CO3Dv2, while layers 0 and 23 need 85\% to 87\% of the tokens on both.
Which layers concentrate their attention is a property of the model and not of one kind of scene.

\FloatBarrier
\section{Hyperparameter Sensitivity}
\label{sec:supp_sparse_hparams}

\input{table/tab_sparse_hparams}

\paragraph{Number of selected frames.}
The accuracy with dense views improves quickly up to $k=3$ and little beyond: the error is 0.047 for $k=1$, 0.037 for $k=3$ and 0.035 for $k=8$ on 7-Scenes, and 0.052, 0.048 and 0.047 on NRGBD (\cref{tab:supp_sparse_hparams}).
The size of the cache does not depend on $k$, because the indexer stores the kept tokens of every frame, and $k$ only sets how many of them a new frame reads.

\paragraph{Share of kept tokens.}
The model is robust to $\rho$.
With $\rho = 0.05$, only 52 patches of a past frame remain, the cache after 1,024 frames shrinks from 6.3 to 1.5\,GiB, and the error rises from 0.037 to 0.042 on 7-Scenes and from 0.048 to 0.050 on NRGBD.
Keeping all tokens ($\rho = 1$) needs 24.5\,GiB and is not consistently more accurate than the default.
The indexer can also be limited to the first two and the last two sparse layers, so that layers 11 and 15, whose teacher attention is the most concentrated among the sparse layers, keep all tokens; this costs 9 to 10\,GiB and gives the accuracy of the model without an indexer.

\FloatBarrier
\section{Limitations}
Our distillation transfers the causal attention of StreamVGGT to Mamba2 layers at a small fraction of the original training cost, but it does not fully remove the drift of a recurrent state on long streams.
Sparse attention recovers part of the lost history, yet the Mamba2 state still compresses hundreds of frames into a fixed size, and we train on clips far shorter than the streams we test on. Camera pose is also less accurate than some state-of-the-art methods.
We therefore see the training recipe itself as the path forward: because a new architecture can be distilled in 9\% of the GPU hours that StreamVGGT needs, future work can iterate quickly over recurrent designs, memory mechanisms and longer training clips, and arrive at models that stay accurate over long streams.

%% file: table/tab_recon_dense.tex
\begin{table}[H]
\centering
\caption{\textbf{3D reconstruction with dense views on 7-Scenes and NRGBD.} One keyframe every 20 frames on 7-Scenes and every 50 frames on NRGBD, which gives 8 to 50 views per sequence. Among online methods, the best result is in \textbf{bold} and the second best is \underline{underlined}.}
\label{tab:supp_recon_dense}
\footnotesize
\setlength{\tabcolsep}{3.4pt}
\resizebox{\linewidth}{!}{%
\begin{tabular}{l c cc cc cc cc cc cc}
\toprule
\multirow{3}{*}[-0.6ex]{\textbf{Method}} & \multirow{3}{*}[-0.6ex]{\textbf{Mode}} & \multicolumn{6}{c}{\textbf{7-Scenes}} & \multicolumn{6}{c}{\textbf{NRGBD}} \\
\cmidrule(lr){3-8} \cmidrule(lr){9-14}
 & & \multicolumn{2}{c}{Acc$\downarrow$} & \multicolumn{2}{c}{Comp$\downarrow$} & \multicolumn{2}{c}{NC$\uparrow$} & \multicolumn{2}{c}{Acc$\downarrow$} & \multicolumn{2}{c}{Comp$\downarrow$} & \multicolumn{2}{c}{NC$\uparrow$} \\
\cmidrule(lr){3-4} \cmidrule(lr){5-6} \cmidrule(lr){7-8} \cmidrule(lr){9-10} \cmidrule(lr){11-12} \cmidrule(lr){13-14}
 & & Mean & Med. & Mean & Med. & Mean & Med. & Mean & Med. & Mean & Med. & Mean & Med. \\
\midrule
VGGT~\cite{wang2025vggt} & Offline & 0.019 & 0.007 & 0.030 & 0.012 & 0.693 & 0.793 & 0.016 & 0.009 & 0.018 & 0.007 & 0.923 & 0.993 \\
$\pi^3$~\cite{wang2025pi3} & Offline & 0.015 & 0.006 & 0.018 & 0.009 & 0.694 & 0.797 & 0.016 & 0.008 & 0.016 & 0.005 & 0.902 & 0.987 \\
FastVGGT~\cite{shen2025fastvggt} & Offline & 0.020 & 0.008 & 0.024 & 0.007 & 0.656 & 0.742 & 0.022 & 0.014 & 0.019 & 0.007 & 0.855 & 0.976 \\
GoToHunt~\cite{zheng2026gotohunt} & Offline & 0.020 & 0.007 & 0.025 & 0.009 & 0.669 & 0.761 & 0.016 & 0.009 & 0.017 & 0.006 & 0.905 & 0.990 \\
ZipMap~\cite{jin2026zipmap} & Offline & 0.018 & 0.009 & 0.032 & 0.011 & 0.681 & 0.780 & 0.018 & 0.011 & 0.019 & 0.007 & 0.893 & 0.988 \\
\midrule
StreamVGGT~\cite{zhuo2025streamvggt} & Online & 0.045 & 0.024 & 0.032 & 0.011 & \textbf{0.678} & \textbf{0.776} & 0.088 & 0.055 & 0.058 & 0.023 & 0.816 & 0.970 \\
CUT3R~\cite{wang2025cut3r} & Online & \underline{0.026} & \underline{0.013} & 0.030 & \underline{0.009} & 0.659 & 0.748 & 0.069 & 0.027 & 0.035 & 0.013 & 0.808 & 0.953 \\
TTT3R~\cite{chen2025ttt3r} & Online & 0.029 & 0.014 & \underline{0.029} & \textbf{0.008} & 0.657 & 0.745 & 0.076 & 0.028 & 0.038 & 0.012 & 0.799 & 0.951 \\
ZipMap~\cite{jin2026zipmap} & Online & \underline{0.026} & \underline{0.013} & 0.030 & \textbf{0.008} & \underline{0.677} & \underline{0.775} & \underline{0.038} & \textbf{0.018} & \underline{0.029} & \underline{0.010} & \textbf{0.851} & \textbf{0.980} \\
LingBot-Map~\cite{chen2026lingbotmap} & Online & \textbf{0.023} & \textbf{0.012} & \textbf{0.027} & \textbf{0.008} & 0.675 & 0.770 & \textbf{0.027} & \textbf{0.018} & \textbf{0.022} & \textbf{0.009} & \textbf{0.851} & \underline{0.976} \\
\textbf{LinearVGGT (Ours)} & Online & 0.037 & 0.020 & 0.030 & 0.010 & 0.665 & 0.758 & 0.048 & \underline{0.024} & 0.030 & 0.012 & \underline{0.827} & 0.971 \\
\bottomrule
\end{tabular}}
\end{table}

%% file: table/tab_vdepth_scale_shift.tex
\begin{table}[H]
\centering
\caption{\textbf{Video depth estimation.} The predicted depth of each sequence is aligned to the ground truth with a scale and a shift. Among online methods, the best result is in \textbf{bold} and the second best is \underline{underlined}.}
\label{tab:supp_vdepth}
\small
\setlength{\tabcolsep}{4pt}
\renewcommand{\arraystretch}{0.92}
\begin{tabular}{l c cc cc cc}
\toprule
 & & \multicolumn{2}{c}{\textbf{Sintel}} & \multicolumn{2}{c}{\textbf{Bonn}} & \multicolumn{2}{c}{\textbf{KITTI}} \\
\cmidrule(lr){3-4} \cmidrule(lr){5-6} \cmidrule(lr){7-8}
\textbf{Method} & \textbf{Mode} & AbsRel$\downarrow$ & $\delta_{1.25}\!\uparrow$ & AbsRel$\downarrow$ & $\delta_{1.25}\!\uparrow$ & AbsRel$\downarrow$ & $\delta_{1.25}\!\uparrow$ \\
\midrule
VGGT~\cite{wang2025vggt}                 & Offline & 0.202             & 0.731             & 0.051             & 0.972             & 0.051             & 0.969 \\
$\pi^3$~\cite{wang2025pi3}               & Offline & 0.221             & 0.736             & 0.044             & 0.975             & 0.036             & 0.986 \\
\midrule
FastVGGT~\cite{shen2025fastvggt}         & Offline & 0.213             & 0.722             & 0.050             & 0.974             & 0.053             & 0.969 \\
GoToHunt~\cite{zheng2026gotohunt}        & Offline & 0.203             & 0.727             & 0.054             & 0.968             & 0.052             & 0.969 \\
\midrule
StreamVGGT~\cite{zhuo2025streamvggt}     & Online  & 0.227             & \underline{0.682} & 0.057             & \underline{0.973} & 0.164             & 0.721 \\
OVGGT~\cite{lu2026ovggt}                 & Online  & 0.236             & 0.670             & 0.057             & 0.971             & 0.120             & 0.855 \\
CUT3R~\cite{wang2025cut3r}               & Online  & 0.499             & 0.546             & 0.066             & 0.963             & 0.117             & 0.866 \\
TTT3R~\cite{chen2025ttt3r}               & Online  & 0.418             & 0.559             & 0.056             & \underline{0.973} & 0.111             & 0.890 \\
ZipMap~\cite{jin2026zipmap}              & Online  & 0.256             & 0.672             & 0.074             & 0.966             & 0.090             & 0.925 \\
LingBot-Map~\cite{chen2026lingbotmap}    & Online  & \textbf{0.221}    & \textbf{0.712}    & \underline{0.054} & 0.966             & \textbf{0.068}    & \textbf{0.962} \\
\textbf{LinearVGGT (Ours)}               & Online  & \underline{0.225} & 0.677             & \textbf{0.047}    & \textbf{0.977}    & \underline{0.074} & \underline{0.947} \\
\bottomrule
\end{tabular}
\end{table}

%% file: table/tab_pose.tex
\begin{table}[H]
\centering
\caption{\textbf{Camera pose estimation.} Absolute trajectory error (ATE), relative translation error (RPE$_t$) and relative rotation error (RPE$_r$, in degrees) after Sim(3) alignment; lower is better. Among online methods, the best result is in \textbf{bold} and the second best is \underline{underlined}.}
\label{tab:supp_pose}
\small
\setlength{\tabcolsep}{4pt}
\begin{tabular}{l c ccc ccc ccc}
\toprule
 & & \multicolumn{3}{c}{\textbf{Sintel}} & \multicolumn{3}{c}{\textbf{TUM-dynamics}} & \multicolumn{3}{c}{\textbf{ScanNet}} \\
\cmidrule(lr){3-5} \cmidrule(lr){6-8} \cmidrule(lr){9-11}
\textbf{Method} & \textbf{Mode} & ATE & RPE$_t$ & RPE$_r$ & ATE & RPE$_t$ & RPE$_r$ & ATE & RPE$_t$ & RPE$_r$ \\
\midrule
VGGT~\cite{wang2025vggt} & Offline & 0.169 & 0.064 & 0.474 & 0.012 & 0.010 & 0.307 & 0.035 & 0.015 & 0.377 \\
$\pi^3$~\cite{wang2025pi3} & Offline & 0.074 & 0.040 & 0.282 & 0.014 & 0.009 & 0.312 & 0.031 & 0.013 & 0.347 \\
FastVGGT~\cite{shen2025fastvggt} & Offline & 0.176 & 0.075 & 0.584 & 0.014 & 0.011 & 0.323 & 0.037 & 0.018 & 0.440 \\
GoToHunt~\cite{zheng2026gotohunt} & Offline & 0.173 & 0.074 & 0.478 & 0.012 & 0.010 & 0.310 & 0.037 & 0.016 & 0.398 \\
\midrule
StreamVGGT~\cite{zhuo2025streamvggt} & Online & 0.232 & 0.094 & 0.714 & \underline{0.027} & \textbf{0.012} & \textbf{0.318} & \underline{0.049} & \underline{0.019} & \underline{0.498} \\
CUT3R~\cite{wang2025cut3r} & Online & 0.213 & 0.066 & 0.621 & 0.046 & 0.015 & 0.473 & 0.099 & 0.022 & 0.600 \\
TTT3R~\cite{chen2025ttt3r} & Online & 0.201 & \textbf{0.063} & \underline{0.617} & 0.028 & \textbf{0.012} & 0.379 & 0.064 & 0.021 & 0.592 \\
ZipMap~\cite{jin2026zipmap} & Online & \underline{0.167} & 0.071 & 0.716 & 0.042 & 0.017 & 0.373 & 0.088 & 0.030 & 0.753 \\
LingBot-Map~\cite{chen2026lingbotmap} & Online & \textbf{0.131} & \underline{0.065} & \textbf{0.473} & \textbf{0.018} & \underline{0.013} & \underline{0.331} & \textbf{0.040} & \textbf{0.018} & \textbf{0.464} \\
\textbf{LinearVGGT (Ours)} & Online & 0.242 & 0.066 & 0.637 & 0.030 & 0.017 & 0.371 & 0.088 & 0.028 & 0.587 \\
\bottomrule
\end{tabular}
\end{table}

%% file: table/tab_per_scene.tex
\begin{table}[H]
\centering
\caption{\textbf{Reconstruction of every test sequence.} Acc$\downarrow$ and Comp$\downarrow$ of StreamVGGT (S) and LinearVGGT (L) with sparse and dense views. The better value of each pair is in \textbf{bold}.}
\label{tab:supp_per_scene}
\footnotesize
\setlength{\tabcolsep}{4.2pt}
\begin{tabular}{l cc cc cc cc}
\toprule
 & \multicolumn{4}{c}{\textbf{Sparse views}} & \multicolumn{4}{c}{\textbf{Dense views}} \\
\cmidrule(lr){2-5} \cmidrule(lr){6-9}
 & \multicolumn{2}{c}{Acc} & \multicolumn{2}{c}{Comp} & \multicolumn{2}{c}{Acc} & \multicolumn{2}{c}{Comp} \\
\cmidrule(lr){2-3} \cmidrule(lr){4-5} \cmidrule(lr){6-7} \cmidrule(lr){8-9}
\textbf{Sequence} & S & L & S & L & S & L & S & L \\
\midrule
\multicolumn{9}{l}{\emph{7-Scenes}} \\
\quad chess/seq-03 & 0.019 & \textbf{0.015} & 0.018 & \textbf{0.015} & 0.021 & \textbf{0.017} & 0.016 & \textbf{0.010} \\
\quad chess/seq-05 & 0.022 & \textbf{0.018} & 0.035 & \textbf{0.020} & 0.021 & \textbf{0.015} & 0.017 & \textbf{0.010} \\
\quad fire/seq-03 & 0.016 & 0.016 & 0.022 & \textbf{0.020} & \textbf{0.010} & 0.012 & \textbf{0.011} & 0.013 \\
\quad fire/seq-04 & 0.058 & \textbf{0.018} & 0.047 & \textbf{0.019} & 0.020 & \textbf{0.009} & 0.014 & \textbf{0.009} \\
\quad heads/seq-01 & 0.047 & \textbf{0.040} & 0.032 & \textbf{0.027} & 0.021 & \textbf{0.015} & 0.014 & \textbf{0.008} \\
\quad office/seq-02 & 0.115 & \textbf{0.075} & 0.116 & \textbf{0.093} & \textbf{0.029} & 0.044 & \textbf{0.022} & 0.027 \\
\quad office/seq-06 & 0.402 & \textbf{0.314} & \textbf{0.194} & 0.218 & \textbf{0.060} & 0.063 & \textbf{0.040} & 0.047 \\
\quad office/seq-07 & 0.203 & \textbf{0.085} & 0.252 & \textbf{0.090} & 0.054 & \textbf{0.037} & 0.031 & \textbf{0.028} \\
\quad office/seq-09 & 0.392 & \textbf{0.078} & 0.354 & \textbf{0.238} & 0.069 & \textbf{0.046} & \textbf{0.037} & 0.075 \\
\quad pumpkin/seq-01 & 0.043 & \textbf{0.021} & 0.049 & \textbf{0.024} & 0.048 & \textbf{0.036} & 0.028 & \textbf{0.023} \\
\quad pumpkin/seq-07 & 0.019 & \textbf{0.016} & \textbf{0.018} & 0.019 & 0.046 & \textbf{0.027} & \textbf{0.020} & 0.023 \\
\quad redkitchen/seq-03 & 0.045 & \textbf{0.029} & 0.035 & \textbf{0.034} & 0.052 & \textbf{0.032} & 0.022 & \textbf{0.019} \\
\quad redkitchen/seq-04 & 0.493 & \textbf{0.120} & 0.409 & \textbf{0.138} & 0.048 & \textbf{0.018} & 0.031 & \textbf{0.014} \\
\quad redkitchen/seq-06 & 0.052 & \textbf{0.034} & 0.054 & \textbf{0.038} & 0.029 & \textbf{0.021} & 0.019 & 0.019 \\
\quad redkitchen/seq-12 & 0.090 & \textbf{0.041} & 0.048 & \textbf{0.034} & 0.055 & \textbf{0.016} & 0.020 & \textbf{0.013} \\
\quad redkitchen/seq-14 & 0.065 & \textbf{0.026} & 0.054 & \textbf{0.033} & 0.039 & \textbf{0.020} & 0.022 & \textbf{0.017} \\
\quad stairs/seq-01 & \textbf{0.184} & 0.186 & 0.220 & \textbf{0.180} & \textbf{0.146} & 0.202 & \textbf{0.118} & 0.120 \\
\quad stairs/seq-04 & 0.064 & \textbf{0.049} & 0.109 & \textbf{0.072} & 0.046 & \textbf{0.038} & 0.088 & \textbf{0.057} \\
\quad \emph{mean} & 0.129 & \textbf{0.066} & 0.115 & \textbf{0.073} & 0.045 & \textbf{0.037} & 0.032 & \textbf{0.030} \\
\midrule
\multicolumn{9}{l}{\emph{NRGBD}} \\
\quad breakfast\_room & \textbf{0.032} & 0.043 & \textbf{0.025} & 0.034 & \textbf{0.019} & 0.033 & \textbf{0.010} & 0.015 \\
\quad complete\_kitchen & 0.077 & \textbf{0.039} & 0.052 & \textbf{0.035} & 0.190 & \textbf{0.036} & 0.124 & \textbf{0.023} \\
\quad green\_room & \textbf{0.056} & 0.077 & \textbf{0.108} & 0.143 & 0.058 & \textbf{0.040} & 0.071 & 0.071 \\
\quad grey\_white\_room & 0.135 & \textbf{0.122} & 0.160 & \textbf{0.056} & 0.203 & \textbf{0.154} & 0.158 & \textbf{0.066} \\
\quad kitchen & \textbf{0.294} & 0.308 & 0.256 & \textbf{0.254} & 0.116 & \textbf{0.067} & 0.075 & \textbf{0.045} \\
\quad morning\_apartment & 0.026 & \textbf{0.018} & 0.028 & \textbf{0.020} & 0.035 & \textbf{0.020} & \textbf{0.017} & 0.018 \\
\quad staircase & \textbf{0.035} & 0.051 & \textbf{0.020} & 0.030 & 0.052 & \textbf{0.045} & 0.022 & \textbf{0.017} \\
\quad thin\_geometry & 0.056 & \textbf{0.019} & 0.027 & \textbf{0.012} & 0.046 & \textbf{0.015} & 0.019 & \textbf{0.007} \\
\quad whiteroom & 0.047 & \textbf{0.025} & 0.032 & \textbf{0.021} & 0.074 & \textbf{0.022} & 0.025 & \textbf{0.011} \\
\quad \emph{mean} & 0.084 & \textbf{0.078} & 0.079 & \textbf{0.067} & 0.088 & \textbf{0.048} & 0.058 & \textbf{0.030} \\
\bottomrule
\end{tabular}
\end{table}

%% file: table/tab_routing.tex
\begin{table}[H]
\centering
\caption{\textbf{Register routing against an oracle.} (a) Agreement with the oracle and recovered share of its attention mass, in percent, over the six sparse layers. (b) The same for the frames chosen by registers in each sparse layer on the 500-frame streams; preference is the mass on the oracle frames relative to three random frames.}
\label{tab:supp_routing}
\small
\setlength{\tabcolsep}{5pt}
\begin{tabular}{l cc cc}
\multicolumn{5}{c}{(a) All sparse layers} \\
\toprule
 & \multicolumn{2}{c}{\textbf{TUM, 90 frames}} & \multicolumn{2}{c}{\textbf{TUM, 500 frames}} \\
\cmidrule(lr){2-3} \cmidrule(lr){4-5}
\textbf{Selected frames} & Agreement & Recovered & Agreement & Recovered \\
\midrule
3 random frames & 12.0 & 78.5 & 3.0 & 66.3 \\
3 most recent frames & 8.0 & 75.0 & 3.3 & 65.8 \\
3 frames chosen by registers (ours) & \textbf{51.6} & \textbf{96.2} & \textbf{43.9} & \textbf{96.4} \\
\bottomrule
\end{tabular}

\vspace{6pt}
\begin{tabular}{l cccccc}
\multicolumn{7}{c}{(b) Registers, per sparse layer} \\
\toprule
\textbf{Layer} & 3 & 7 & 11 & 15 & 19 & 23 \\
\midrule
Preference ($\times$) & 1.18 & 1.14 & 2.51 & 1.97 & 1.22 & 1.07 \\
Agreement & 36 & 20 & 100 & 74 & 19 & 14 \\
Recovered & 96.1 & 95.1 & 100 & 96.4 & 91.2 & 96.0 \\
\bottomrule
\end{tabular}
\end{table}

%% file: table/tab_sparse_hparams.tex
\begin{table}[H]
\centering
\caption{\textbf{Sensitivity of the sparse attention layers to their hyperparameters.} Every row changes one setting of the final model at test time, without retraining; the default setting is repeated in each group. Acc$\downarrow$ with dense views and the size of the sparse-layer caches after 1,024 frames.}
\label{tab:supp_sparse_hparams}
\small
\setlength{\tabcolsep}{5pt}
\begin{tabular}{l cc cc c}
\toprule
 & \multicolumn{2}{c}{\textbf{7-Scenes} Acc} & \multicolumn{2}{c}{\textbf{NRGBD} Acc} & \textbf{Cache} \\
\cmidrule(lr){2-3} \cmidrule(lr){4-5}
\textbf{Setting} & Mean & Med. & Mean & Med. & (GiB) \\
\midrule
\multicolumn{6}{l}{\emph{Selected frames $k$}} \\
\quad $k = 1$ & 0.047 & 0.024 & 0.052 & 0.025 & 6.3 \\
\quad $k = 2$ & 0.041 & 0.022 & 0.053 & 0.026 & 6.3 \\
\quad $k = 3$ (default) & 0.037 & 0.020 & 0.048 & 0.024 & 6.3 \\
\quad $k = 5$ & 0.038 & 0.019 & 0.046 & 0.022 & 6.3 \\
\quad $k = 8$ & 0.035 & 0.019 & 0.047 & 0.023 & 6.3 \\
\midrule
\multicolumn{6}{l}{\emph{Token budget $\rho$}} \\
\quad $\rho = 0.05$ & 0.042 & 0.025 & 0.050 & 0.026 & 1.5 \\
\quad $\rho = 0.1$ & 0.040 & 0.022 & 0.050 & 0.026 & 2.7 \\
\quad $\rho = 1/4$ (default) & 0.037 & 0.020 & 0.048 & 0.024 & 6.3 \\
\quad $\rho = 0.5$ & 0.039 & 0.021 & 0.046 & 0.023 & 12.4 \\
\quad $\rho = 1$ (no indexer) & 0.042 & 0.021 & 0.046 & 0.023 & 24.5 \\
\midrule
\multicolumn{6}{l}{\emph{Indexer only in layers 3, 7, 19, 23}} \\
\quad $\rho = 0.1$ & 0.042 & 0.021 & 0.046 & 0.023 & 9.9 \\
\quad $\rho = 0.05$ & 0.043 & 0.021 & 0.045 & 0.023 & 9.1 \\
\bottomrule
\end{tabular}
\end{table}